\documentclass[10pt,twocolumn,letterpaper]{article}
\usepackage{silence}
\usepackage{microtype}
\usepackage{graphicx}
\usepackage{booktabs} 

\usepackage{iccv}
\usepackage{times}
\usepackage{epsfig}
\usepackage{graphicx}

\usepackage{mathtools, amssymb, amsmath}
\usepackage[utf8]{inputenc} 
\usepackage[T1]{fontenc}    
\usepackage{url}            
\usepackage{amsfonts}       
\usepackage{nicefrac}       
\usepackage{xcolor}         
\usepackage{colortbl}
\usepackage{subcaption}
\usepackage{enumitem}
\usepackage{multirow}
\usepackage{capt-of}
\usepackage{wrapfig}
\usepackage{makecell}
\usepackage{bm}
\usepackage{arydshln}
\usepackage{placeins}
\usepackage{bmpsize}

\usepackage{algorithm}
\usepackage{algorithmic}
\usepackage[pagebackref=true,breaklinks=true,colorlinks,bookmarks=false]{hyperref}

\iccvfinalcopy 

\def\iccvPaperID{9870} 

\ificcvfinal\fi
\definecolor{orange-red}{rgb}{1.0, 0.27, 0.0}
\definecolor{Red}{rgb}{1,0,0}
\definecolor{darkred}{rgb}{0.9, 0.17, 0.31}

\newcommand{\supp}[1]{{\color{darkred}{#1}}} 
\newcommand{\tref}[1]{{\color{darkred}{Table~\ref{#1}}}}
\newcommand{\fref}[1]{{\color{darkred}{Figure~\ref{#1}}}}
\newcommand{\sref}[1]{{\color{darkred}{Section~\ref{#1}}}}
\newcommand{\aref}[1]{{\color{darkred}{Algorithm~\ref{#1}}}}
\begin{document}

\title{Text-Conditioned Sampling Framework for\\ Text-to-Image Generation with Masked Generative Models}
\author{Jaewoong Lee$^*$\\
KAIST\\
Republic of Korea\\
{\tt\small hello3196@kaist.ac.kr}
\and
Sangwon Jang$^*$\\
Yonsei University\\
Republic of Korea\\
{\tt\small agwmon@gmail.com}
\and
Jaehyeong Jo, Jaehong Yoon\\
KAIST\\
Republic of Korea\\
{\tt\small \{harryjo97, jaehong.yoon\}@kaist.ac.kr}
\and
Yunji Kim, Jin-Hwa Kim, Jung-Woo Ha\\
NAVER AI Lab\\
Republic of Korea\\
{\tt\small \{yunji.kim, jinhwa.kim, jungwoo.ha\}@navercorp.com}
\and
Sung Ju Hwang\\
KAIST\\
Republic of Korea\\
{\tt\small sjhwang82@kaist.ac.kr}
}

\maketitle
\ificcvfinal\thispagestyle{empty}\fi
\def\thefootnote{*}\footnotetext{Equal contribution.}
\def\thefootnote{}\footnotetext{Preliminary work.}
\begin{abstract}
Token-based masked generative models are gaining popularity for their fast inference time with parallel decoding. While recent token-based approaches achieve competitive performance to diffusion-based models, their generation performance is still suboptimal as they sample multiple tokens simultaneously without considering the dependence among them. We empirically investigate this problem and propose a learnable sampling model, Text-Conditioned Token Selection (TCTS), to select optimal tokens via localized supervision with text information. TCTS improves not only the image quality but also the semantic alignment of the generated images with the given texts. To further improve the image quality, we introduce a cohesive sampling strategy, Frequency Adaptive Sampling (FAS), to each group of tokens divided according to the self-attention maps. We validate the efficacy of TCTS combined with FAS with various generative tasks, demonstrating that it significantly outperforms the baselines in image-text alignment and image quality. Our text-conditioned sampling framework further reduces the original inference time by more than $\mathit{50\%}$ without modifying the original generative model.
\end{abstract}

\vspace{-4mm}
\section{Introduction} 

\begin{figure*}[t]
    \centering
    \includegraphics[width=1.0\linewidth]{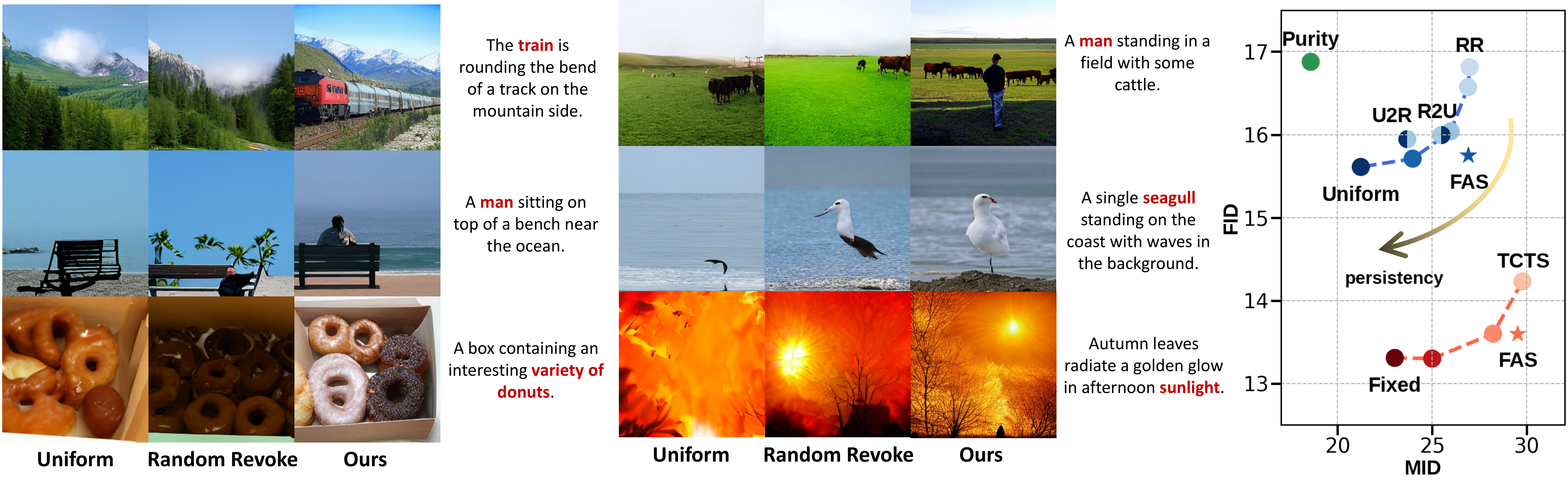} 
\vspace{-0.3in}
    \caption{\small \textbf{Generated samples on MS-COCO dataset and evaluation graph of various sampling methods showing their trade-off.} Uniform sampling is a fixed strategy with notably poor text alignment compared to other methods (FID-40K: 15.61, MID-L: 21.23). Random revoke sampling is a revocable strategy with improved text alignment (FID-40K: 16.81, MID-L: 26.98). Ours is TCTS combined with FAS, where both the image quality and the text alignment are significantly better compared to those of baselines (FID-40K: 13.6, MID-L: 29.5). Metrics are measured on all 40K images with their corresponding single caption. The classifier-free guidance scale was fixed at 5 for all sampling methods.}
    \label{Samples} 
\vspace{-0.177in}
\end{figure*}


In the flood of generative AI systems for vision domains, text-conditional image generation~\cite{nichol2021glide, saharia2022photorealistic, yu2022scaling} is coming to the fore in recent years. Although many recent works have achieved success in synthesizing high-quality images~\cite{ho2022cascaded, sauer2022stylegan} with plausible class-alignment in class-conditional cases~\cite{casanova2021instance, dhariwal2021diffusion}, text-to-image (T2I) generation is more challenging since generating visual outputs that are semantically aligned with input texts is a nontrivial problem. We can roughly categorize the works on text-to-image generation into transformer-based \emph{autoregressive} (AR)~\cite{gafni2022make, ramesh2021zero, saharia2022photorealistic} and \emph{diffusion-based}~\cite{ho2020denoising, ramesh2022hierarchical} approaches. Along with the advancement of language models, AR models using transformers have shown impressive performance in text-to-image generation. Despite their success, they suffer from the problem of \emph{unidirectional bias}, which is undesirable for image generation, and crucially, the sampling process requires over 10 times as much time compared to existing models. Another line of work is diffusion-based methods that aim to generate images by iteratively denoising noisy samples. In particular, several continuous diffusion models \cite{fan2022frido, rombach2022high} have achieved outstanding performance and reduced computational cost. Yet, they require excessive sampling steps to obtain high-quality images during the inference.

Recently, a new family of generative models, called \emph{token-based diffusion model}~\cite{chang2022maskgit, lee2022draft, yu2021vector}, has emerged as an alternative to tackle the problem of text-to-image generation. Token-based diffusion models quantize the latent features into tokens and apply categorical corruption process in discrete state spaces~\cite{austin2021d3pm}, while conventional diffusion models use Gaussian noise in continuous space. Among various discrete diffusion methods, mask-based diffusion, similar to the absorbing state diffusion used in \cite{austin2021d3pm}, is mostly used. Compared to existing AR models, this token-based approach is advantageous for speeding up the generation process via simultaneously sampling multiple tokens. Despite the limitation on the reconstruction capacity, these token-based diffusion models significantly outperform the competitors in terms of FID scores, even with fewer sampling steps compared to the continuous diffusion models ~\cite{chang2023muse, tang2022improved}. 

However, sampling multiple tokens at once often leads to inconsistency throughout a generated image~\cite{tang2022improved}. For each location, the generator outputs a probability distribution that are coherent with each other. However, there is no guarantee that every single token sampled from the distributions will perfectly align with one another. In other words, it is still possible for the generator to sample incompatible tokens regardless of the generator's capability, leading to potentially nonsensical outputs~\cite{lezama2022improved,tang2022improved}.
This results in a trade-off between the sampling steps and generation quality. Reducing the sampling steps leads to faster generation but results in performance degradation due to a large number of simultaneously sampled tokens. Especially, this problem further stands out in the text-to-image generation tasks since the distribution of text-aligned images is more restricted than the distribution of unconditioned or class-conditioned images.

To address these limitations, we propose a novel sampling approach that refines the images during the diffusion process based on the text condition, which we refer to as the \emph{Text Conditioned Token Selection} (TCTS). TCTS is a sampling strategy that can mask out and resample previously sampled tokens, which we refer to as \emph{revocable}. To find the tokens that do not align with the given text condition, 
we propose to train a model that selects tokens to be masked out and re-generated, to be well-aligned with the given text. Combining this approach with the revocable sampling scheme, TCTS can generate high-quality images with improved text-alignment in even fewer sampling steps compared to the naive generative model, as shown in \fref{Samples}. We further introduce \emph{Frequency Adaptive Sampling} (FAS) to solve the over-simplification problem that occurs when applying revocable methods for relatively longer steps. FAS leverages the self-attention map of the images and applies a mixed sampling method, which we call \emph{persistent sampling} to prevent the issue. We summarize our contributions as follows:
\begin{itemize}
    \item We experimentally show that the revocable sampling strategies are crucial for the trade-off between the text-alignment and the image quality and provide in-depth analysis compared to previous fixed sampling methods.
    \item We propose a novel revocable sampling method based on a learnable token selection model that significantly improves the text alignment as well as the image quality even with fewer sampling steps, without the need of retraining the generator model.
    \item Moreover, we propose a novel sampling method based on the self-attention map that can be combined with TCTS to solve the over-simplification problem.
\end{itemize}

\section{Related work}
Various works have tackled text-to-image generation tasks, and the majority of them are based on Generative Adversarial Networks (GANs) \cite{tao2020df, xu2018attngan, zhang2017stackgan, zhang2018stackgan++,zhou2022lafite2, zhou2022towards, zhu2019dm}. However, computing directly from the pixel space of an image is challenging, and generating high-resolution images requires substantial computation due to the large number of pixels involved. Therefore, a two-stage approach is often used: models first tokenize the images into a sequence of codes, and then predict the tokenized codes.

\vspace{-0.1in}
\paragraph{Autoregressive models}
Following the success of Transformer~\cite{vaswani2017attention} and GPT~\cite{radford2018improving}, Autoregressive (AR) models~\cite{esser2021taming, ramesh2021zero, razavi2019generating, yu2022scaling} proposed to tokenize images and pose the text-to-image generation problem as a problem of translating textual descriptions into image tokens, much like machine translation of one language into a different language. With the advancement of image tokenization and language models, this approach has achieved very high performance. However, even with the help of this tokenizing stage~, there remains a large number of tokens to predict, resulting in a significant delay in the generation time due to their autoregressive nature. Additionally, errors are accumulated through the irrevocable process, with undesirable unidirectional bias.

\vspace{-0.1in}
\paragraph{Latent diffusion models}
Recently, diffusion models~\cite{ho2020denoising, song2020score} introduced a new paradigm for text-to-image generation~\cite{saharia2022photorealistic}. They generate high-quality images by progressively denoising noise, but require iterative steps and a large amount of computation. Two-stage approaches~\cite{ fan2022frido, rombach2022high} propose to perform diffusion in latent space successfully reducing computation while maintaining performance. However, most of them still require about 50 inference steps to show plausible performance, and since denoising is applied globally to the entire image during the inference process, locally correcting the image requires complicated processes~\cite{rombach2022high}.

\vspace{-0.1in}
\paragraph{Token-based diffusion models} 
Token-based diffusion models~\cite{chang2023muse, gu2022vector, lee2022draft} have impressive performance in generating complex scenes from the text. Unlike conventional diffusion models, where the Gaussian noise is applied in a continuous space, token-based diffusion models quantize latent features into tokens and apply a categorical corruption process in a discrete state space~\cite{austin2021d3pm}. Among various discrete diffusion methods, mask-based diffusion is most commonly used due to its attractive properties; they have the advantage of sampling multiple tokens simultaneously, resulting in fewer sampling steps and faster generation time than the latent diffusion model. However, selecting multiple tokens at once leads to inconsistency throughout the image which is often called a joint distribution issue. Moreover, tokens are heavily influenced by the spatially adjacent tokens that have already been modified so that a misaligned token could disperse the errors throughout the image. To tackle this issue, some works~\cite{chang2022maskgit, lee2022draft} modified the sampling strategy while other works \cite{lezama2022improved} introduced new models to find the misaligned tokens.

In particular, Lezama \etal.~\cite{lezama2022improved} proposed a learnable model that removes the misaligned tokens at each step. While most existing works randomly select the tokens to discard, this work examines all sampled tokens. It outputs the score of each location where the score indicates whether the corresponding token is under the real distribution. After the predicting and sampling process of the generator, the model selects the locations according to the scores of the corresponding tokens.

\section{Text-conditioned token selection}
The diffusion process of the masked image generative models starts with a completely masked image, and samples the tokens of $x_t$ for the whole image to predict $\hat{x}_0$ while re-masking a portion of them to obtain the denoised sample $x_{t-1}$. Existing masked image generative models use uniform or confidence-based sampling strategy to determine the locations to re-mask: \emph{Uniform sampling}~\cite{gu2022vector, lee2022draft} randomly selects the locations of the tokens to keep among the predicted tokens at all locations, while confidence-based strategy, such as \emph{purity sampling}~\cite{tang2022improved}, selects the locations based on the confidence score which is defined as follows:
\begin{align}
    \texttt{CS}(i,t) = \max\limits_{j=1,\cdots,K} p(x_0^i=j|x_t^i) ,
\end{align}
where $t$ is the step number, $K$ is the size of the codebook, and $i$ is the location of the token. Intuitively, a high confidence score means the generator is relatively certain that a token should be sampled at the corresponding location. By keeping these confident tokens while discarding the ambiguous ones, purity sampling shows good performance in practice~\cite{chang2023muse}.

However, these strategies are non-revocable, \ie, \emph{fixed}, which means tokens that are once sampled cannot be revised afterward, even if they do not match the given text or do not align with other tokens. Once the misaligned tokens are fixed during the process, errors accumulate and spread throughout the image. We observe that fixed strategies negatively affect the text alignment as visualized in \fref{Samples}.

\begin{figure}[t!]
\centering
    \includegraphics[width=1.0\linewidth]{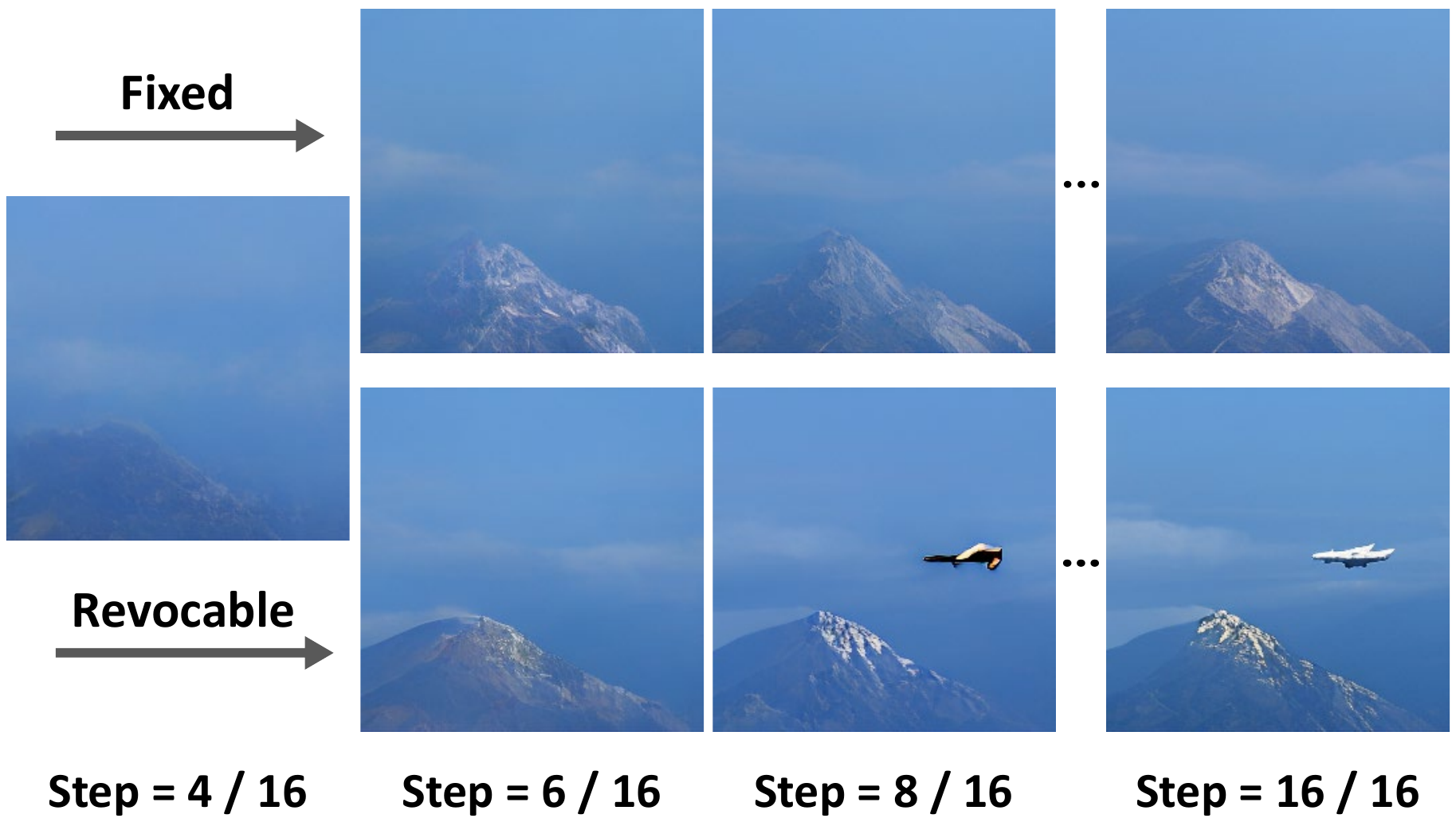} 
\vspace{-0.25in}
     \caption{\small \textbf{Reconstructed images during diffusion using the fixed and revocable method.} Only the revocable method is able to edit the tokens to generate a plane according to the text "\emph{A view of the end of an airplane in the sky over mountains.}".}
    \label{revocable_plane}
\vspace{-0.15in}
\end{figure}

\subsection{Random revoke sampling strategy}\label{3.1}
In order to address this issue, we first introduce a new sampling strategy called \emph{random revoke sampling} and use it as a baseline for our final method. It is similar to uniform sampling in a way that they both determine the locations by sampling from a uniform distribution. However, while uniform sampling is a fixed strategy, random revoke sampling selects the locations from the whole image space regardless of the previously fixed locations, making the process revocable. It randomly selects a certain number of tokens to preserve for the next step and revokes the previously fixed ones. The revocable feature can better align the images to the given text information, especially in complex scenes. 

To be specific, as shown in the top row of \fref{revocable_plane}, fixed strategies have limitations in recovering from errors or generating diverse images due to their inability to regenerate incorrect tokens or modify missing essential objects or attributes. In contrast, revocable strategies enhance the text alignment by providing an opportunity to correct invalidly generated tokens, as shown in the bottom row of \fref{revocable_plane}. Since revocable strategies have no constraints on which tokens to remove or fix, they have the advantage of mitigating error accumulation. 

\begin{figure}[t]
\centering
    \includegraphics[width=1.0\linewidth]{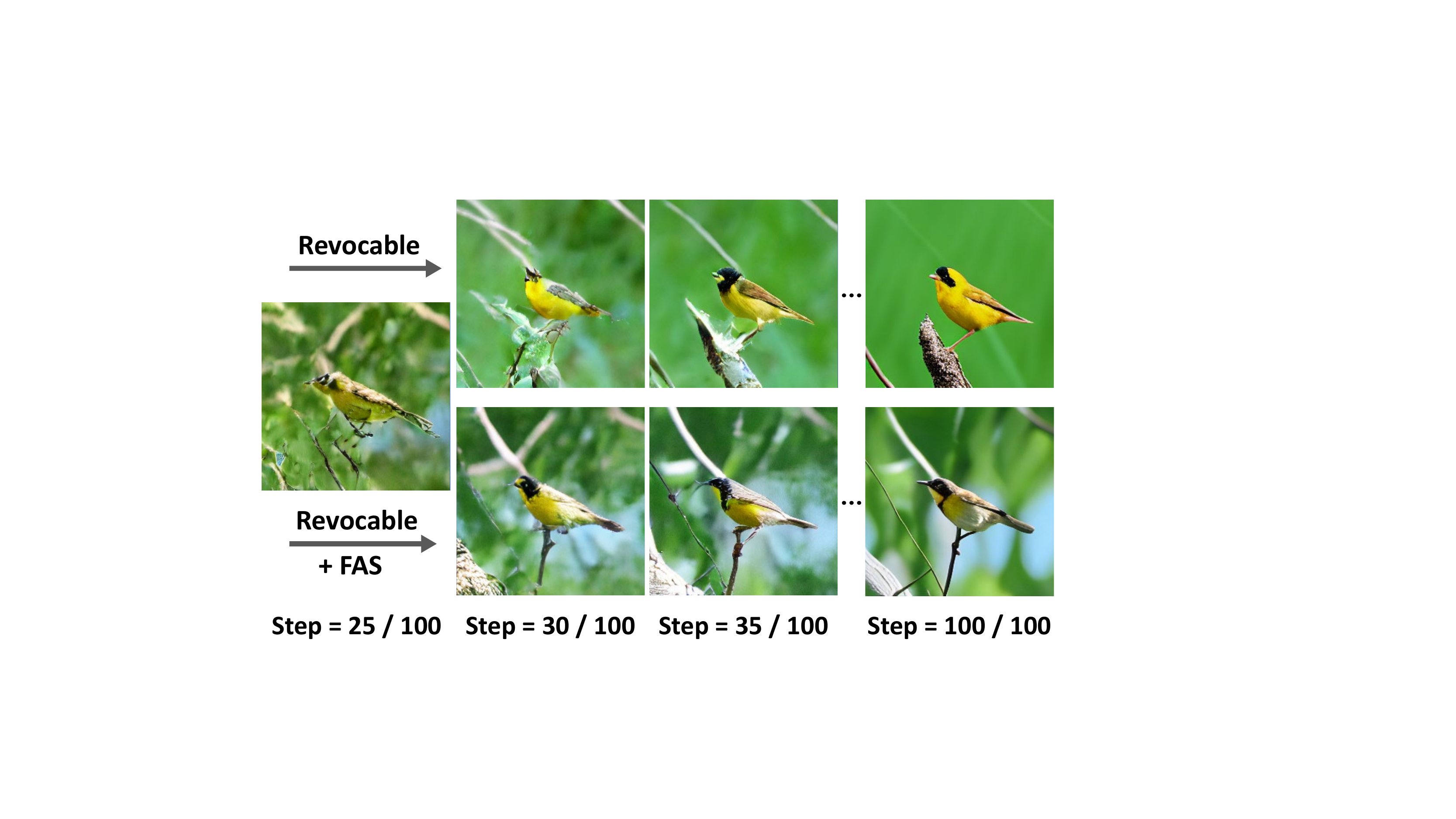}
\vspace{-0.25in}
     \caption{\small \textbf{Reconstructed images during diffusion using the revocable method with and without FAS.} The backgrounds of the images in the top row are over-simplified while our proposed FAS prevents this, as shown in the bottom row. The text is "\emph{This small bird is greyish in color with flecks of yellow on the back and breast, and a bit of white on the belly.}".}
    \label{fig:FAS_sample}
\vspace{-0.15in}
\end{figure}
However, random revoke sampling suffers from two main issues that must be addressed. The first issue is the instability resulting from randomly discarding previously sampled tokens which worked as a given condition of the step. It can lower the quality of the outputs as in \fref{COCO}. 

The second issue, namely the \emph{over-simplification}, arises when it is used in the generation process involving longer steps. The low-frequency parts of the output images, such as backgrounds, tend to be over-simplified compared to fixed methods, which generate more realistic backgrounds. The over-simplified samples can be seen in \fref{fig:FAS_sample}, and it has a significant impact on the FID metric as in \tref{COCO} and \ref{CUB}.
We observe that this is due to excessive repetition of resampling over a large number of steps, while a fixed schedule does not resample any tokens. More opportunities for resampling make the tokens converge towards safer and simpler patterns. As shown in the top row of \fref{fig:FAS_sample}, the high-frequency areas with rich visual details do not become over-simplified even after multiple resampling processes, but the low-frequency areas become less diverse and more simplified in patterns quickly due to excessive resampling.

To confirm this trade-off between text alignment and image quality, we define a \emph{persistent sampling} as an interpolation between uniform sampling and random revoke sampling. In details, a persistent weight boosts the probability of keeping the tokens that were sampled in the previous step. While random revoke sampling assigns the same probability to all the locations, persistent sampling multiplies the persistent weight to the probability of previously kept locations. Further details are shown in \aref{algo:algorithm_cs}. If the persistent weight $w$ is 1, all the tokens would have the same probability to be fixed, such as random revoke sampling. If $w$ is sufficiently large, the previously sampled tokens would be kept until the end, such as uniform sampling. In the graph in \fref{Samples}, a trade-off can be seen that as the sampling strategy gets closer to the random revoke sampling, text alignment gets better, while the image quality gets worse. We further analyze the effect of these sampling methods in \sref{further}.
\begin{figure}[t]
\vspace{-0.15in}
\begin{minipage}{\linewidth}
\centering
\begin{algorithm}[H]
    \small
    \caption{Persistent Sampling}
	\label{algo:algorithm_cs}
        \textbf{Input:} $k_t$: number of tokens sampled at step $t$, $w$: persistent weight, $N$: number of all tokens
	\begin{algorithmic}[1]
            \STATE $x_T \leftarrow [$[MASK]$]_N$
	    \FOR {$t = T, T-1, ..., 1$}
                \STATE $\hat{x}_0 = G_\theta(x_t,c)$
                \STATE $A_t = \{i|x_t^i=$[MASK]$\}$
                \STATE $m = k_T + k_{T-1} + ... + k_t$
                \STATE  $\mathcal{U}_t(I=i|i \in A_t):\mathcal{U}_t(I=i|i \in A_t^C)= 1:w$
                \STATE $\rightarrow$ uniform distribution + persistent weight
                \STATE $i_1, i_2, ..., i_m \sim \mathcal{U}(I) \leftarrow $ sample without replacement
                \STATE $x_t \leftarrow[$[MASK]$]_N$
                \FOR {$i = i_1, i_2, ..., i_m$}
                    \STATE $x^i_t \leftarrow \hat{x}^i_0$
                \ENDFOR
            \ENDFOR
            \STATE \textbf{Return:} Generated image $x_0$
        \end{algorithmic}
    \end{algorithm}
\end{minipage}
\vspace{-0.1in}
\end{figure}



\begin{figure*}[t]
\centering
    \includegraphics[width=1.0\linewidth]{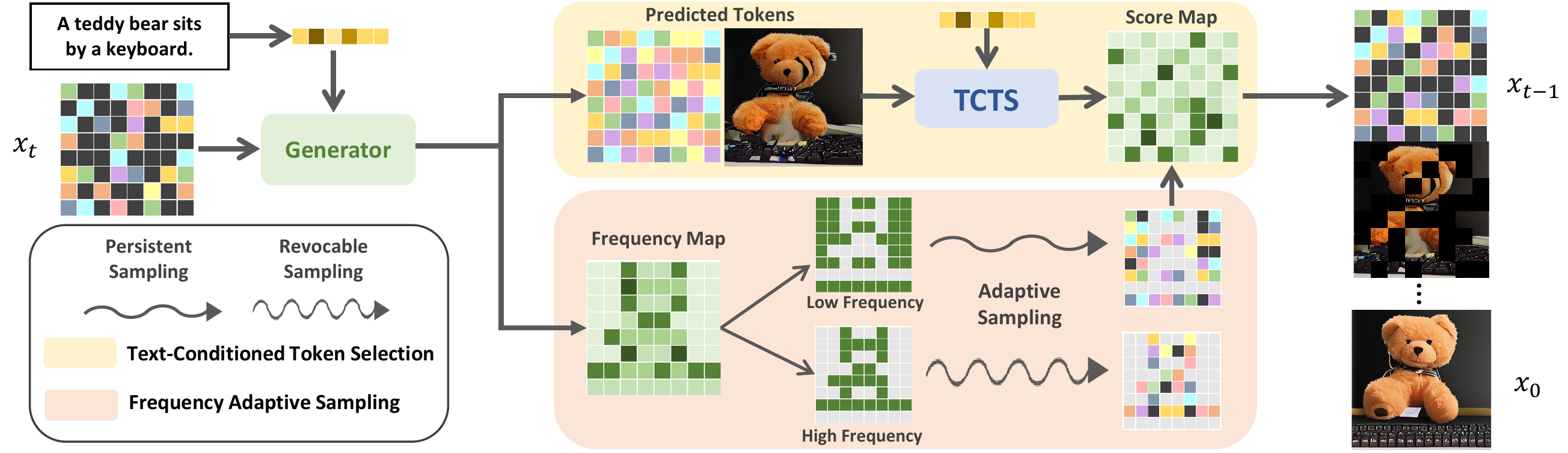}
\vspace{-0.2in}
    \caption{\small \textbf{Overall generation framework of proposed TCTS and FAS.} After the generator predicts the tokens, TCTS exploits the text condition to detect misaligned tokens and outputs the score map. Meanwhile, FAS splits the tokens according to the frequency using the self-attention map from the generator, performing revocable sampling to high-frequency split and persistent sampling to low-frequency split. The adaptive sampling predicts $\hat{x}_0$ and decide a few of the locations to mask according to $x_t$. The token maps produced by FAS is combined with the score map to predict $x_{t-1}$. After iterative process, our model removes all the masks and generates $x_0$.}
    \label{fig:concept}
\vspace{-0.0in}
\end{figure*}

\subsection{Text-conditioned token selection}
We propose a learnable model called \emph{Text-Conditioned Token Selection} (TCTS). TCTS is trained to output a score map to detect tokens that are not under the real distribution for the given text condition, and selects the tokens that are not aligned with the text or other tokens in the image. By masking the selected tokens at each step, the diffusion network receives high-quality text-aligned tokens as the input to sample the next set of tokens which alleviates the error accumulation and joint distribution issue of the previous sampling strategies. Moreover, since TCTS is trained to discriminate the well-aligned tokens, it detours the instability problem as mentioned earlier, unlike other revocable methods. From the experimental results in \sref{Exp_sec}, we observe that TCTS enhances the text alignment without compromising the image quality.

\paragraph{Frequency adaptive sampling (FAS)}\label{FAS}
The simplest way to solve the over-simplification problem in longer steps is to use the persistent sampling to alleviate the excessive resampling of tokens. However, as previously mentioned, this method may result in a slight compromise in text alignment and limit the opportunity to correct the image through resampling. To further improve this trade-off, we propose a new method called \emph{Frequency Adaptive Sampling (FAS)}, which can be applied to TCTS.

%
\begin{table}[t]
    \centering
    \resizebox{1.0\linewidth}{!}{
    \renewcommand{\arraystretch}{1.2}
    \renewcommand{\tabcolsep}{5pt}
    \begin{tabular}{clcccc}
    \toprule
    Step & Method & ~MID-L $\uparrow$ & SOA-I $\uparrow$ & ~~CLIP-S $\uparrow$ &  ~~~FID-30K $\downarrow$\\
    \midrule
    \multirow{4}{*}{16} & Purity & 11.02 & 72.38 & 0.2474 & 19.20 \\
    & Uniform & 17.94 & 74.80 & 0.2500 & 16.17 \\
    & RR & 23.60 & 78.79 & 0.2526 & 17.10 \\
    \cmidrule{2-6}
    & \textbf{TCTS + FAS} & \textbf{26.72} & \textbf{79.52} & \textbf{0.2559} & \textbf{14.45} \\
    \midrule
    \multirow{4}{*}{25} & Purity & 16.84 & 75.21 & 0.2487 & 18.39 \\
    & Uniform & 22.27 & 77.08 & 0.2524 & 15.91 \\
    & RR & 26.77 & \textbf{81.10} & 0.2543 & 18.43 \\
    \cmidrule{2-6}
    & \textbf{TCTS + FAS} & \textbf{27.79} & 80.87 & \textbf{0.2563} & \textbf{15.39} \\
    \bottomrule
    \end{tabular}}
\vspace{-0.1in}
    \captionof{table}{\small \textbf{Quantitative evaluation of sampling methods on MS-COCO dataset.} The ground truth MID on MS-COCO image-caption pairs is $54.73$. The classifier-free guidance scale was fixed at 5 for all sampling methods.}
    \label{COCO}
\vspace{-0.15in}
\end{table}
FAS is a method that utilizes the generator's self-attention map to limit resampling only in the low-frequency areas of the image. In \cite{hong2022selfa}, it is known that the generator's self-attention layer contains frequency information of the image, which is also observed in token-based diffusion models. We utilized this to distinguish the frequency areas of the image without additional operations, and applied persistent sampling only to the low-frequency areas using persistent weight. As shown in the bottom row of \fref{fig:FAS_sample}, revocable sampling along with FAS method allows repetitive resampling in areas that require rich visual details, while limiting the number of resampling in relatively simple areas preventing over-simplification. Detailed algorithm is in the \supp{supplementary A}. The overall framework of our model, including both TCTS and FAS, is illustrated in \fref{fig:concept}.

\begin{table}[t]
    \centering
    \resizebox{1\linewidth}{!}{
    \renewcommand{\arraystretch}{1.0}
    \renewcommand{\tabcolsep}{8pt}
    \begin{tabular}{clccc}
    \toprule
    Step & Method & ~MID-L $\uparrow$ & ~CLIP-S $\uparrow$ & ~FID $\downarrow$\\
    \midrule
    \multirow{4}{*}{16} & Purity & -24.21 & 0.2410 & 15.21 \\
    & Uniform & -25.60 & 0.2404 & 16.57 \\  
    & RR & -25.03 & 0.2371 & 17.38 \\
    \cmidrule{2-5}
    & \textbf{TCTS + FAS} & \textbf{-19.88} & \textbf{0.246
    }& \textbf{12.35} \\
    \midrule
    \multirow{4}{*}{25} & Purity & -21.26 & 0.2384 & \textbf{12.60} \\
    & Uniform & -23.04 & 0.2396 & 13.02 \\
    & RR & -23.29 & 0.2364 & 14.53 \\
    \cmidrule{2-5}
    & \textbf{TCTS + FAS} & \textbf{-18.31} & \textbf{0.2409}
    & 13.67 \\
    \bottomrule
    \end{tabular}}
\vspace{-0.1in}
    \caption{\small \textbf{Quantitative evaluation of sampling methods on CUB dataset.} The ground truth MID on CUB image-caption pairs is 15.85. We omit SOA in this table as CUB dataset consists of only one bird per photo, diminishing the metric's significance.}
    \label{CUB}
\vspace{-0.1in}
\end{table}
 
\section{Experiments}\label{Exp_sec}
To validate the efficacy of our token selection framework, we modify the transformer from \cite{tang2022improved} to contain 310M parameters for MS-COCO dataset and 83M parameters for CUB dataset. 
We extract text features from CLIP ViT-B/32~\cite{radford2021learning} for a text-conditioned generation. In the training process of TCTS, we freeze the generator's parameters and independently train our model, making it applicable to other types of token-based diffusion models. We use the binary cross-entropy loss for the objective and validate our method and baselines on MS-COCO~\cite{lin2014microsoft} and CUB~\cite{he2019fine} datasets by training them for 200 epochs. We leverage classifier-free guidance~\cite{ho2022classifier} by stochastic sampling of the guidance strength to improve the generation quality. We further provide the details of the architecture, hyperparameters and additional analysis on the use of classifier-free guidance in the \supp{supplementary A}. 

\paragraph{Metrics}
Since FID~\cite{heusel2017gans} metric is known to be problematic for its inability to exactly represent the image quality~\cite{chang2023muse} and unable to consider the text alignment, we additionally use Mutual Information Divergence (MID)~\cite{kim2022mutual} to evaluate the text-alignment of generated images which enables sample-wise evaluation and outperforms previous metrics in human Likert-scale judgment correlations. In particular, since MID responds more sensitively to images generated with foiled captions~\cite{shekhar2017foil}, it is more appropriate for analyzing text alignment with complex and challenging captions. 
Moreover, we use the CLIP score~\cite{hessel2021clips} and SOA-I~\cite{hinz2020semantic}, which are widely used in text-to-image synthesis. Note that the CLIP score is calculated with ViT-L/14 throughout the experiments.

\subsection{Text-to-image synthesis}
\fref{Samples} Left and Middle visualizes the generated images using improved VQ diffusion~\cite{tang2022improved} with varying sampling methods. Generated examples from uniform token sampling are poorly aligned with the given texts, and often include erratic partitions of the objects since the predicted tokens selected at the same step cannot appropriately reflect the change of others (\eg, \emph{smashed donuts, and imperfect chair and seagull}).
Random revoke (RR) sampling, which can re-update previously sampled tokens, seems to improve alignment with text by editing out nonsensical regions and iteratively considering the entire scene at each generation step.
Yet, RR selects tokens in a completely random manner without any constraints, which results in suboptimal generation quality and alignment with the texts (\eg, \emph{disappeared man and train}).
On the other hand, ours mitigates the problem from RR by selecting text-conditioned tokens at the sampling phase and successfully generates high-quality images which contain a clear semantic connection to the given text captions. 


\begin{table}[t]
    \centering
    \resizebox{1.0\linewidth}{!}{
    \renewcommand{\arraystretch}{1.1}
    \renewcommand{\tabcolsep}{4pt}
    \begin{tabular}{lcccc}
    \toprule
    Model & ~~MID-L $\uparrow$ & ~~MID-B $\uparrow$ & ~SOA-I $\uparrow$ &  ~FID-30K $\downarrow$\\
    \midrule
    GLIDE \cite{nichol2021glide} & ~~~1.03 & ~~1.00 & - & 32.08\\
    AttnGAN \cite{xu2018attngan} & -65.20 & ~-8.90 & 39.01 & 29.15\\
    DM-GAN \cite{lee2022draft} & -44.66 & ~~~3.51 & 48.03 & 22.90\\
    DF-GAN \cite{tao2020df} & -58.75 & -15.21 & - & 31.75 \\
    VQ-Diffusion \cite{yu2021vector} & -19.63 & ~~~5.77 & - & 13.13\\
    LAFITE \cite{zhou2022towards} & ~~~6.26 & ~35.17 & 74.78 & \textbf{~~8.03} \\
    \midrule
    \textbf{TCTS + FAS} & ~\textbf{26.02} & ~\textbf{38.98} & \textbf{79.52} & ~~9.75 \\
    \bottomrule
    \end{tabular}}
    \vspace{-0.05in}
    \captionof{table}{\small \textbf{Comparison of the proposed method with recent text-to-image generative models on MS-COCO dataset.} MID-L and MID-B are calculated with ViT-L/14 and ViT-B/32 each. The ground truth MID-L and MID-B on MS-COCO image-caption pairs is 54.73, 41.57 each. Ours are evaluated with 16-step setting and we adopt classifier-free guidance from 3 to 5.}
    \label{Metric}
\vspace{0.in}
\end{table}

This phenomenon can be quantitatively seen in \tref{COCO}. 16-step generation of random revoke sampling shows better text alignment than even 25-step generation of other fixed sampling methods. We can also observe the trade-off between CLIP score and FID by comparing each method to their 25-step versions. Our model outperforms other baselines in most of the metrics especially when comparing MID-L to fixed sampling methods such as Purity and Uniform. Since our model can make corrections during the generation process, it requires fewer steps to match the performance of our baseline sampling methods, see \sref{Inference_Time} for further analysis. We also present the results of our experiment on the CUB dataset with fewer parameters in \tref{CUB}, which demonstrates satisfactory performance. However, the fact that the CUB dataset contains only single object per image resulted in a tendency that is slightly different from what we mentioned earlier. In \tref{Metric}, we compared the performance of our method with other text-to-image generative models. Our model highly exceeds other models in text alignment metrics, especially in MID-L and closely approximate the MID-B value of the ground truth images. More samples generated by our model is in the \supp{supplementary B}.

\begin{figure}[t]
\centering
\vspace{-0.025in}
    \includegraphics[width=1.0\linewidth]{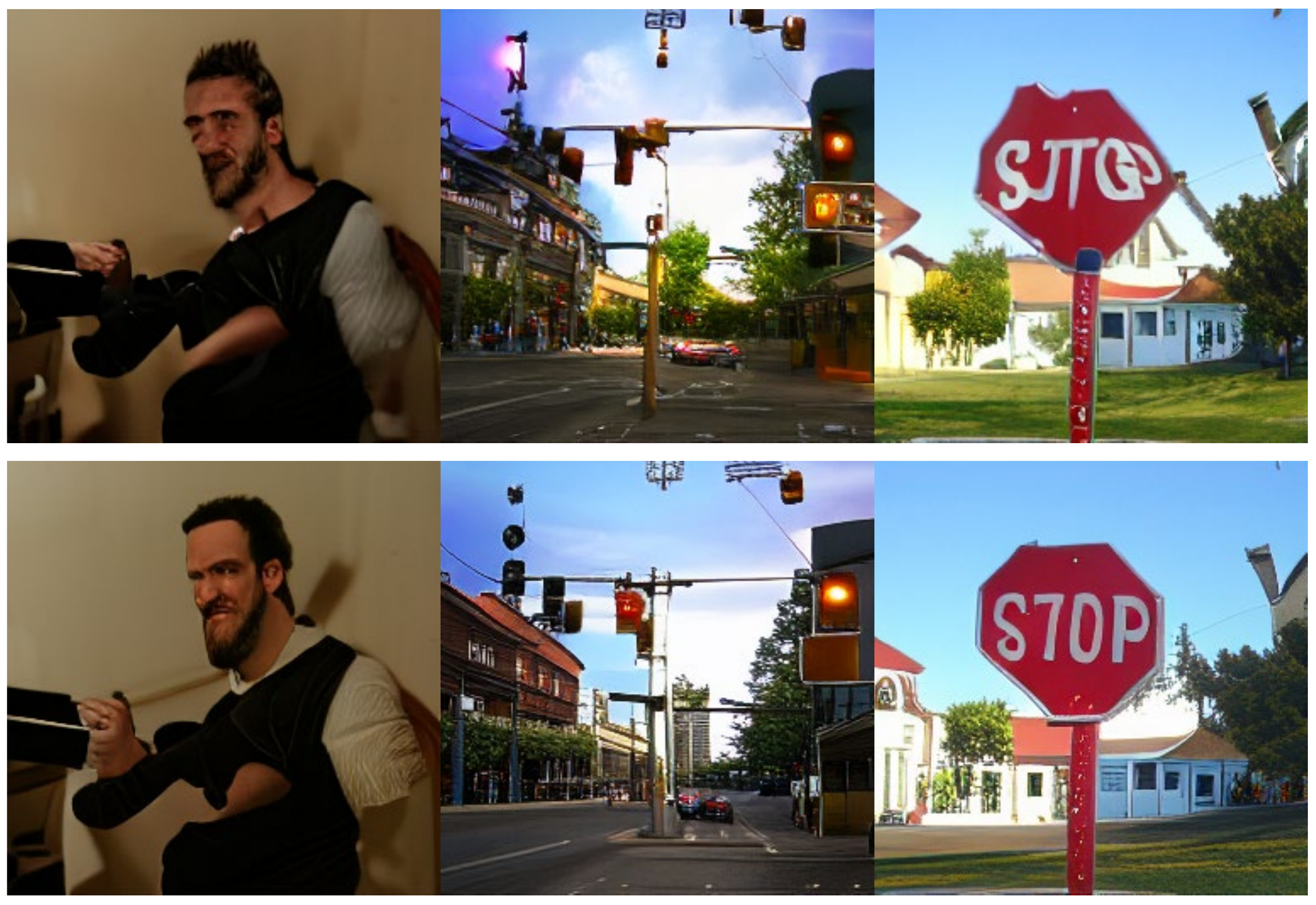}
\vspace{-0.3in}
    \caption{\small \textbf{Results of image refinement using TCTS.} \textbf{Top}: Original samples,  \textbf{Bottom}: Refined images for 8-steps with TCTS.}
    \label{fig:img_refine}
\end{figure}

\begin{table}[t]
\vspace{-0.15in}
    \centering
    \resizebox{0.7\linewidth}{!}{
    \renewcommand{\arraystretch}{1.0}
    \renewcommand{\tabcolsep}{10pt}
    \begin{tabular}{lccc}
    \toprule
    Method & ~~MID-L $\uparrow$ & ~FID $\downarrow$\\
    \midrule
    Original & 12.10 & \textbf{16.44} \\
    \midrule
    RR& 13.64 & 17.18 \\
    TCTS & \textbf{15.64} & 16.92 \\
    \midrule
    \midrule
    Original & ~~4.54 & 19.13 \\
    \midrule
    Random masked & ~~6.06 & 19.12 \\
    TCTS masked & ~~\textbf{6.47} & \textbf{18.64} \\
    \bottomrule
    \end{tabular}}
\vspace{-0.1in}
    \caption{\small \textbf{Qualitative evaluation of the refined images.} \textbf{Top}: Refinement with additional revision steps. \textbf{Bottom}: Refinement with masking lowest-scoring tokens.}
    \label{tab:img_refine}
\vspace{-0.15in}
\end{table}  
\subsection{Image refinement}\label{4.2}
The advantage of the masked image generative model is that it enables fast local refinement. We conduct image refinement in two separate methods, inspired by \cite{lee2022draft} and \cite{lezama2022improved}. First, we use TCTS to apply additional revision steps to images generated with uniform sampling. The overall image quality is improved by adding additional refinement steps as shown in \fref{fig:img_refine}. In order to demonstrate the image refinement performance of our TCTS, we also measure the performance of the image refinement with random revoke sampling. While RR improves does improve sample quality, additional refinement steps increase the FID score, caused by a similar effect to increasing the generation step.

We additionally perform the experiment by masking 60\% lowest-scoring tokens with TCTS, and generate the images with uniform sampling. We observe that all metrics of the refined images outperform those of the original images. In \tref{tab:img_refine}, we compare them with samples refined after randomly masking the tokens without TCTS. Due to the page limit, we describe further details of the two refinement methods and visualize more samples in the \supp{supplementary B}.

\subsection{Mask-free object editing}
Since the masked image generative model is capable of local refinement, image editing without manual masking is possible simply by randomly masking a part of the image and resampling it with the new text condition. However, this method requires a low masking ratio and many resampling steps to maintain the overall structure of the image. This can result in significant changes to unnecessary parts or, as mentioned earlier, over-simplification issues. Additionally, even with many steps, editing large objects with small masks can be more challenging due to the significant influence of surrounding tokens rather than the new text condition. 

\begin{figure}[t!]
\centering
    \includegraphics[width=1.0\linewidth]{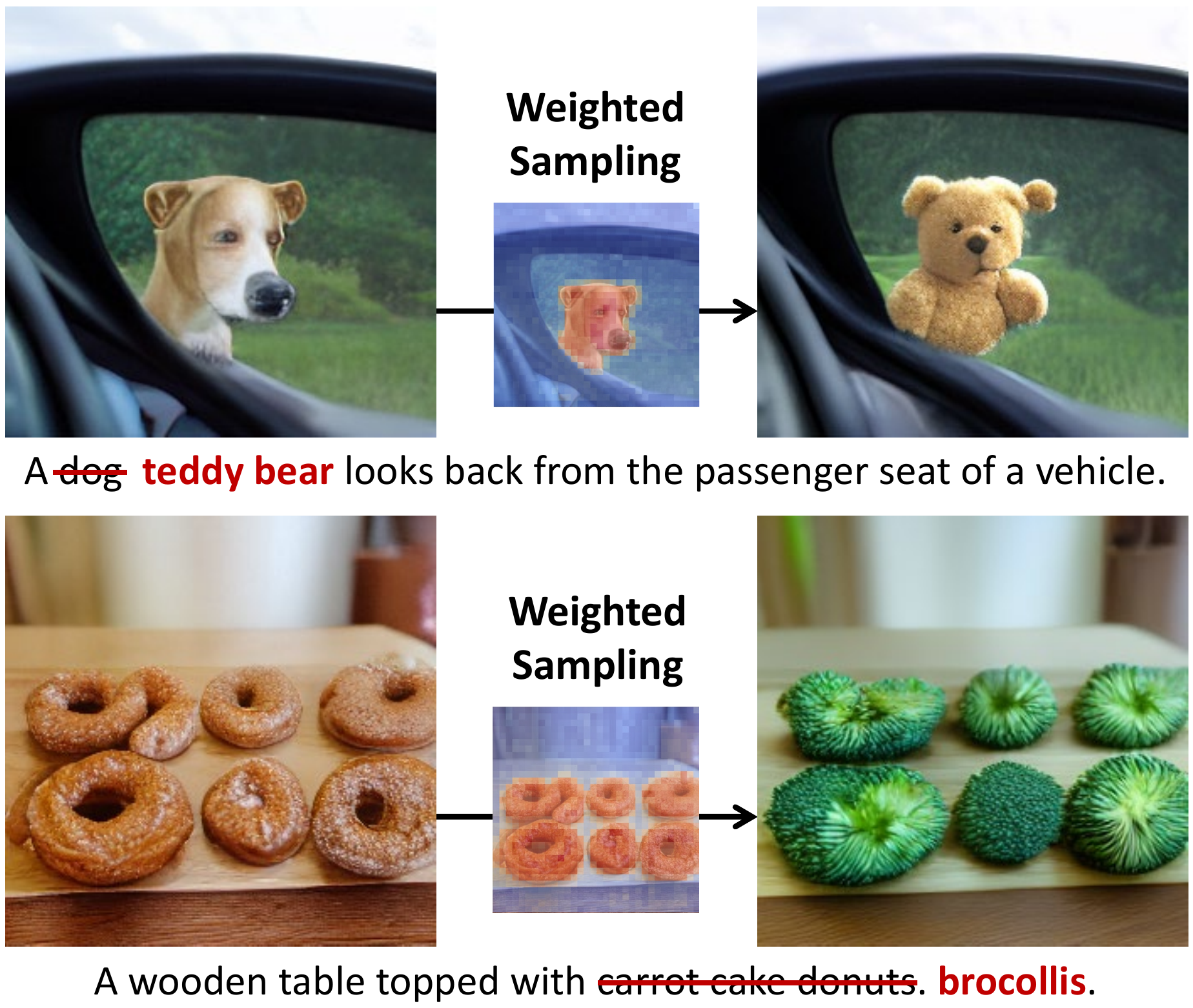}
\vspace{-0.2in}
    \caption{\small \textbf{Examples of mask-free editing samples with cross-attention map.} The cross-attention map is multiplied to the score map of TCTS to perform weighted sampling.}
    \label{fig:mask-free} 
\vspace{-0.in}
\end{figure}

Motivated by the operation of self-attention maps in frequency adaptive sampling, we leverage a cross-attention map corresponding to the word of the object that is to be changed, giving weights to resample tokens so that the corresponding locations can be resampled. Then, it enables efficient image editing with fewer steps and makes it easier to edit larger objects. Although it is similar to DiffEdit~\cite{couarion2022diffedit}, ours operates in a more straightforward manner without additional computations, thanks to the capability of local refinement of the masked generative model. In \fref{fig:mask-free}, we edit the object by adding 25\% noise in only 10-steps. With our method, we can edit more quickly with fewer editing steps, which further does not suffer from the over-simplification problem. More samples are in the \supp{supplementary B}.

\subsection{High-resolution image synthesis}
\begin{figure}
\centering
    \includegraphics[width=1.0\linewidth]{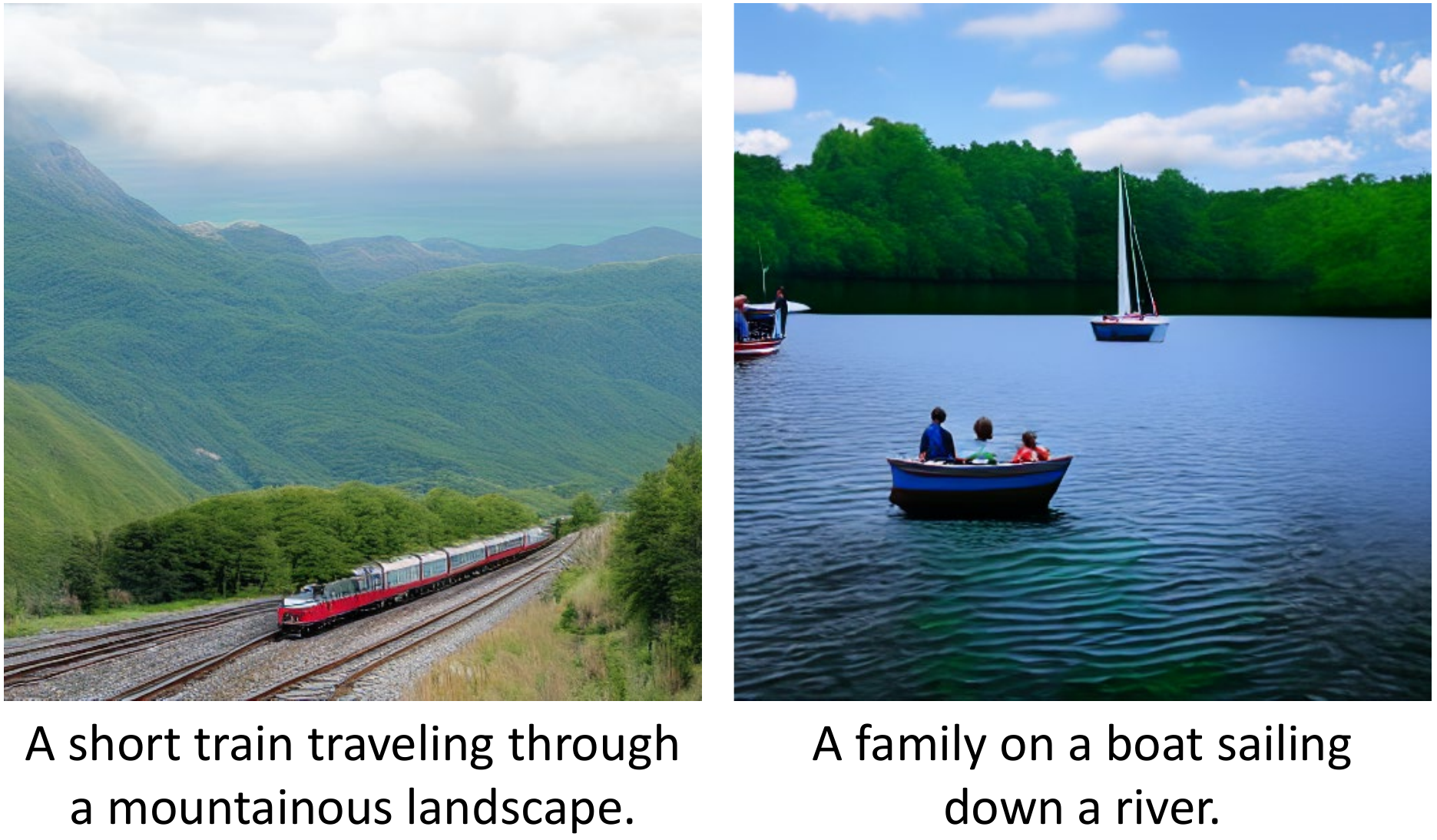}
\vspace{-0.2in}
     \caption{\small \textbf{Examples of high resolution ($\mathbf{512^2}$) samples.} \textbf{Left}: Image generated with all mask tokens, \textbf{Right}: Image generated with TCTS as a super-resolution unit.}
    \label{fig:high-resolution}
\vspace{-0.1in}
\end{figure}

Continuous diffusion models that use encoders and decoders, such as masked image generative models, can generate larger images that are not in the training set. Similar to Bond-Taylor \etal.~\cite{bond2022unleashing}, we divide the tokens into subsets according to the model input size, individually pass them through the model, and then spatially aggregate tokens to synthesize a high-resolution image. In this way, we are able to generate more realistic and high-resolution samples with the same TCTS model without additional training. Additionally, we propose a new method to generate high-resolution images only with low-resolution TCTS. To be specific, we first generate a small-size image with TCTS and upsample the token map in bicubic mode to the desired size. Then we divide the high-resolution token map into overlapping small-size sections and refine all the sections several times. Then, we can generate high-resolution images with the low-resolution TCTS where we visualize the generated samples in \fref{fig:high-resolution}.

\subsection{Further analysis}\label{further}

\paragraph{Importance of early-stage sampling strategy} 
As mentioned in \sref{3.1}, by controlling the persistent weight, we can interpolate the two baseline sampling methods: uniform sampling and random revoke sampling. We designed various experiments to find out the impact of the sampling method with respect to the time steps. We switched the sampling methods during the time steps, from uniform sampling to random revoke sampling (U2R) and vice versa (R2U). In the rightmost plot in \fref{Samples}, we can observe that U2R shows better text-alignment performance than R2U.

These results suggest that the early stage sampling, in which the masked ratio is high, substantially influences the text alignment of the final image, which is similarly observed in continuous diffusion models~\cite{balagi2022ediff}. Generating the whole image and then processing it through a few revision steps like \cite{lee2022draft} might help the image quality but can not conspicuously enhance the essential text alignment. Therefore, in order to obtain an image that aligns well with the text, sampling must be done carefully in the early stage when not many tokens are generated. This analysis is also shown in the result that the revocable strategy exhibits better performance by aggressively resampling tokens several times. In particular, since weak generators with poor generation performance suffer more from the joint distribution issues, giving a sufficient number of recovery chances to discard tokens and draw new ones at the early stage is the key to the desired text alignment.


\paragraph{Text-conditioned sampling} 
\begin{figure}[t!]
\centering
    \includegraphics[width=1.0\linewidth]{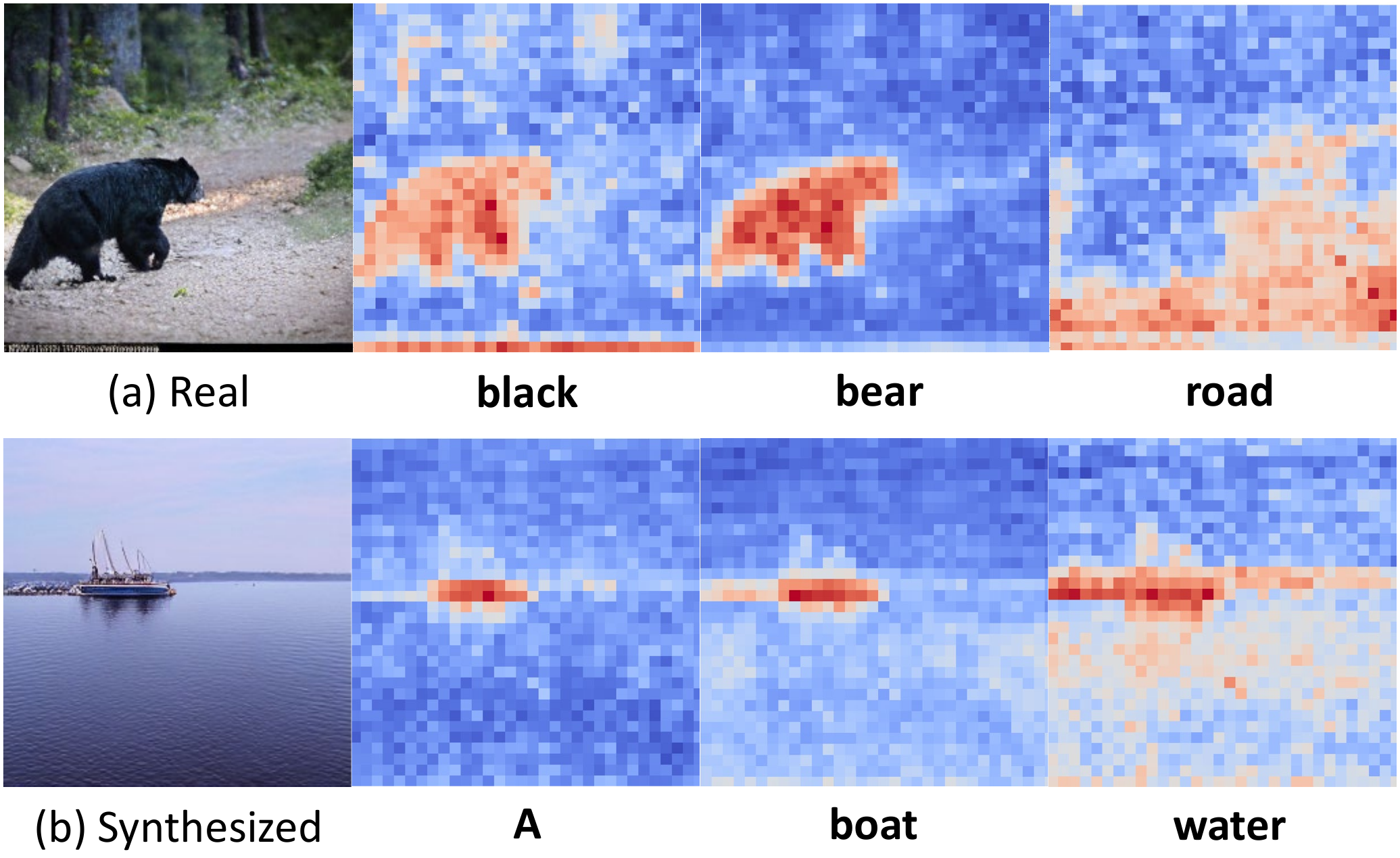}
\vspace{-0.25in}
     \caption{\small \textbf{Visualization of the averaged cross-attention maps for each word from TCTS.} \textbf{Top}: Image from COCO dataset: "\emph{A large black bear walking down a dirty road.}".  \textbf{Bottom}: Synthesized image: "\emph{A small boat in a narrow body of water.}".}
    \label{fig:attn}
\vspace{-0.1in}
\end{figure}
The experiments show that TCTS plays a crucial role in enhancing text alignment in image generation. Although we did not utilize any text-specific loss during training, the model still effectively processes the text condition and achieves optimal performance. To verify the accuracy of TCTS's attention towards text-related parts of the image, we visualized the average cross-attention layer for each word in \fref{fig:attn}. The results demonstrate TCTS's ability to attend to each word in both real and synthesized images. By helping the selection of relevant tokens during the generating stage, TCTS can boost model performance without the requirement of additional complex text-related loss.


\begin{table}[t]
\centering
    \resizebox{\linewidth}{!}{
    \renewcommand{\arraystretch}{1.1}
    \renewcommand{\tabcolsep}{4pt}
    \begin{tabular}{lcccc}
    \toprule
    Method & ~MID-L $\uparrow$ & ~SOA-I $\uparrow$ & ~~~CLIP-S $\uparrow$ &  ~~~FID-30K $\downarrow$\\
    \midrule
    RR & \textbf{26.77} & \textbf{81.10} & \textbf{0.2543} & 18.43 \\
    RR + FAS & 26.32 & 80.78 & 0.2539 & 16.01 \\
    RR + pw & 24.23 & 79.14 & 0.2534 & \textbf{15.96} \\
    \midrule
    TCTS & \textbf{27.82} & \textbf{80.93} & \textbf{0.2565} & 16.69 \\
    TCTS + FAS & 27.79 & 80.87 & 0.2563 & 15.39 \\
    TCTS + pw & 25.12 & 79.51 & 0.2556 & \textbf{15.12} \\
    \bottomrule
    \end{tabular}}
\vspace{-0.1in}
    \caption{\small \textbf{Ablation study of using FAS and persistent weight in revocable methods on MS-COCO dataset.} All models were evaluated on 25-step setting. The ground truth MID-L on MS-COCO image-caption pairs is $54.73$.}
    \label{FAS_ab}
\vspace{-0.1in}
\end{table}

\paragraph{Ablation study of FAS}
As mentioned in \sref{FAS}, FAS decides whether to apply persistent weight or not to each location. If the whole image is considered low-frequency, FAS would apply the weight to every location, which is the same as the persistent sampling. On the other hand, if the whole image is considered high-frequency, combining FAS would not apply the weight anywhere, which does not change the sampling strategy. To further analyze the effect of FAS, we evaluated it on two revocable methods: random revoke sampling and TCTS. Attaching FAS to a base sampling method can be regarded as a mixture of two methods: the base revocable sampling method and persistent sampling method. \tref{FAS_ab} clearly shows this relation of FAS and persistent weight. While FAS-attached methods do not show the best performance, the performance of them is always in close proximity to the better one of the two comparing methods of each. FAS-attached methods balances well between text-alignment and image quality without any additional training.

\begin{table}[t]
    \centering
    \resizebox{1.0\linewidth}{!}{
    \renewcommand{\arraystretch}{1.0}
    \renewcommand{\tabcolsep}{8pt}
    \begin{tabular}{clccc}
    \toprule
    Step & Model & MID-L $\uparrow$ & FID-30K $\downarrow$ & Time \\
    \midrule
    \multirow{2}{*}{50} & Uniform & 24.04 & 16.84 & \multirow{2}{*}{$\times$1} \\
    & RR & 26.55 & 21.12 &  \\
    \midrule
    16 & TCTS + FAS & \textbf{26.72} & \textbf{14.45} & \textbf{$\times$0.45} \\
    \bottomrule
    \end{tabular}}
\vspace{-0.1in}
    \captionof{table}{\small \textbf{Inference time relative to MID/FID of our baselines and models on MS-COCO dataset.} We expressed the inference time of our models as a multiple of other baselines. Our model outspeeds other baselines while performing better in both MID-L and FID-30K. Further analysis is in the \supp{supplementary C}.}
    \label{Inference}
\vspace{-0.1in}
\end{table}
\paragraph{Inference time}\label{Inference_Time}
Our TCTS and FAS compute additional operations at each generation step, resulting in marginally increased inference time compared to other token selection baselines. However, regarding the text alignment and the image quality, the relative inference time is considerably decreased. In \tref{Inference}, our model outperforms 50-step image generation of baseline sampling methods within only 16 steps in MID-L and FID-30K, which means that our model can synthesize better samples with only $\times0.45$ time.


\section{Conclusion}
This paper examines which factors in the masked diffusion process impact the output images and cause the trade-off between image quality and text alignment. We empirically find that the revocable sampling significantly improves the text alignment yet degrades the quality of the generated images.
To tackle the problem, We propose a simple token sampling strategy, coined text-conditioned token selection, that combines learnable and revocable methods, pushing the boundary of the trade-off between image quality and text alignment without the necessity to update the large-scale pre-trained generator.
We find that collaborative sampling in a persistent and revocable manner surprisingly alleviates over-simplification issues in the generated backgrounds. Our proposed method can be utilized as an image refinement tool for various generative tasks, which is remarkably fast to generate high-quality images within much fewer steps.




{\small
\bibliographystyle{ieee_fullname}
\bibliography{egbib}
}

\newpage
\appendix
\onecolumn
\section{Implementation Details}





\subsection{Guidance sampling training}
\begin{figure*}[ht]
\centering
\vspace{0.2in}
    \includegraphics[width=1.0\linewidth]{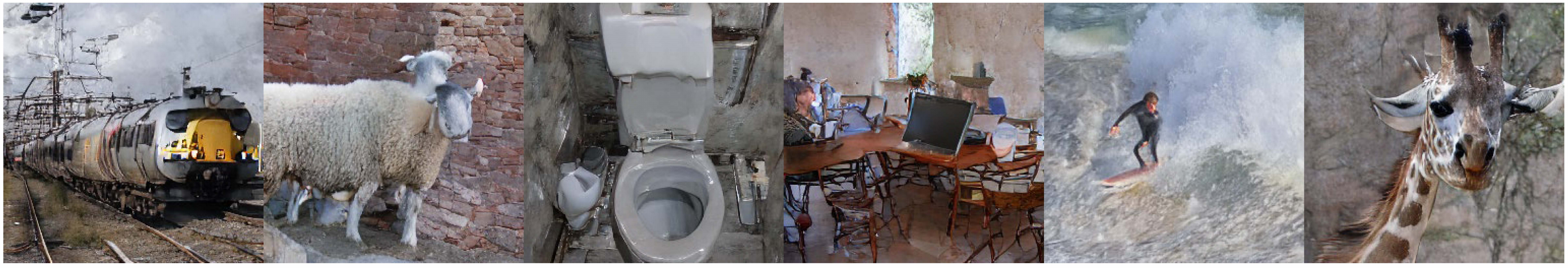}
\vspace{-0.2in}
    \caption{\small \textbf{Generated samples with TCTS trained with no classifier-free guidance (FID-30K: 27.13, MID-L: 12.34).}}
    \label{fig:appn_guidance}
\vspace{-0in}
\end{figure*}
Classifier-free guidance~\cite{ho2022classifier} is a key factor for the image quality of text-to-image diffusion models~\cite{nichol2021glide, saharia2022photorealistic}, and also has been successfully applied to the token-based models~\cite{chang2023muse, tang2022improved}. Since classifier-free guidance is used at inference time, it must also be used during the training process. However, determining the guidance scale during training is a difficult problem. We found out that either overly high or low guidance scale can deteriorate the training process of TCTS. The training procedure of TCTS starts with masking random tokens of a real image. Then, a fixed generator reconstructs the image and TCTS is trained to find the originally masked locations. However, since high guidance scale boosts the reconstruction capacity, TCTS suffers from finding the masked locations and tends to output a smooth distribution. On the other hand, a lowercase with poor reconstruction performance provides the model with diverse and easy samples, making the learning process faster and more stable. However, since a high guidance scale is used during actual inference, the model exhibits very low performance. The samples are in \fref{fig:appn_guidance}. Therefore, we stochastically sample the guidance scale in the training procedure as a regularization of the difficulty of the task. This guidance scale controlling method stabilizes training, improves performance, and enables various guidance scale settings at inference time.

\subsection{Frequency Adaptive Sampling}
\begin{figure*}[h]
\centering
\vspace{0.2in}
    \includegraphics[width=1.0\linewidth]{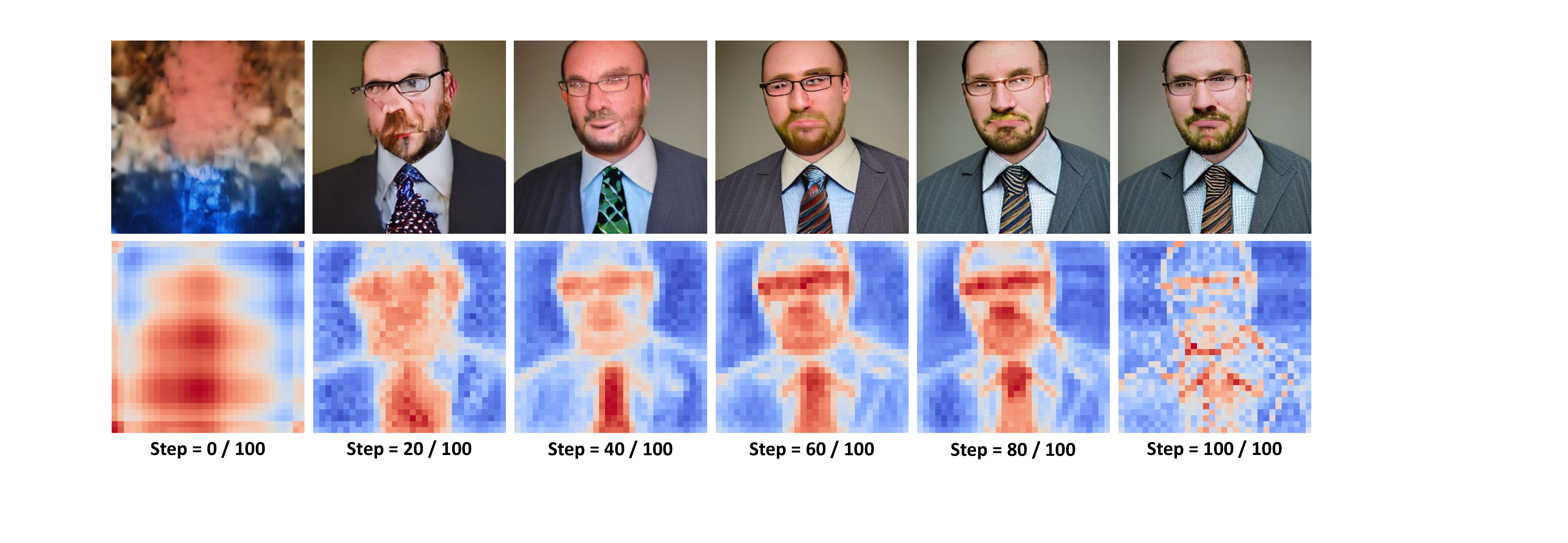}
\vspace{-0.2in}
    \caption{\small \textbf{Visualization of self-attention map}. \textbf{Top}: Reconstructed images in each step. \textbf{Bottom}: Visualization of self-attention maps for each step.}
    \label{fig:appn_vis_self}
\vspace{-0in}
\end{figure*}
\begin{figure*}[ht]
\centering
\vspace{0.2in}
    \includegraphics[width=1.0\linewidth]{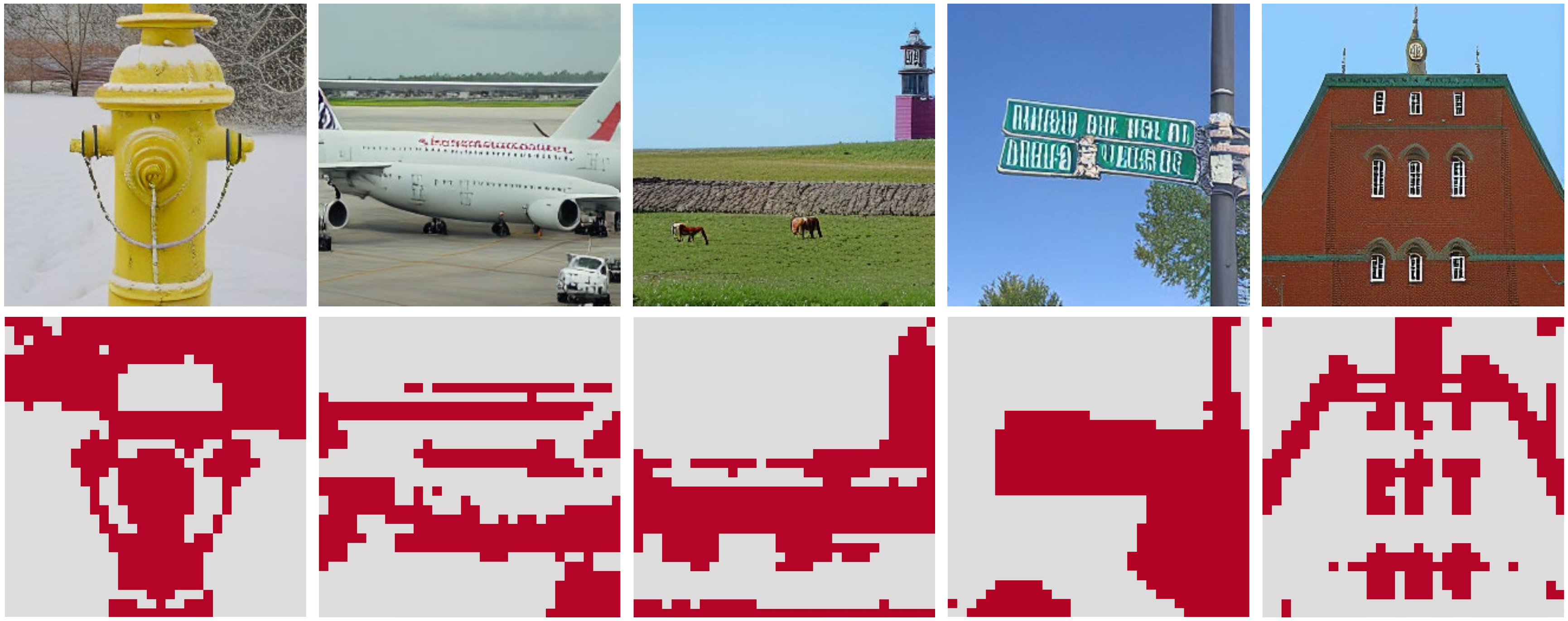}
\vspace{-0.2in}
    \caption{\small \textbf{Visualization of self-attention map with threshold} \textbf{Top}: Synthesized images. \textbf{Bottom}: Visualization our self-attention masks. ($\phi$ = 0.5)}
    \label{fig:appn_self_ths}
\vspace{-0in}
\end{figure*}
The area where highly detailed information is obtained in \fref{fig:appn_vis_self} has large values in the high-frequency range. It was observed that as the generating process progressed, the values gradually decreased in all areas, especially in the simplified areas. Changes were first observed in the areas that were simplified as a priority. In our FAS method, we divided this area using a threshold, and the visualization of this is shown in \fref{fig:appn_self_ths}. This is similar to the object mask or frequency mask that can be found in \cite{hong2022selfa}. The detailed algorithm is in \aref{algo:algorithm_fas}.
\begin{figure}[ht]
\vspace{-0.15in}
\begin{minipage}{\linewidth}
\centering
\begin{algorithm}[H]
    \small
    \caption{Frequency Adaptive Sampling}
	\label{algo:algorithm_fas}
        \textbf{Input:} $w$: persistent weight, $map_{sa}$: self-attention map, $\phi$: self-attention threshold \\
        $G_\theta$: Generator, $D_\gamma$: TCTS model, $c$: text condition embedding
	\begin{algorithmic}[1]
            \STATE $\hat{x}_0$, $map_{sa}$ = $G_\theta(x_t, c)$ \\
            \STATE $map_{tc}$ = $D_\gamma(\hat{x}_0, c) \leftarrow$ TCTS probability map   
            \STATE $A_t = \{i|x_t^i=$[MASK]$\}$ 
            \STATE $L_t = \{i|map_{sa}^i < \phi\} \leftarrow$ Low frequency location
            \STATE $a = 1 + (w-1)\times(n(A^C) \div N)$ 
            \FOR {$i \in  A_t^C \cap L_t$} 
                \STATE $map_{tc}^i = map_{tc}^i\times w$
            \ENDFOR
            \FOR {$i \in  A_t$}
                \STATE $map_{tc}^i = map_{tc}^i\times a$
            \ENDFOR
           \STATE \textbf{Return:} $map_{tc}$
        \end{algorithmic}
    \end{algorithm}
\end{minipage}
\vspace{-0.1in}
\end{figure}

\noindent In addition, unlike \cite{hong2022selfa}, the self-attention map here uses sigmoid instead of softmax on the values before the softmax calculation in the original transformer. 
\begin{align}
    map_{sa}^{(h)} = sigmoid(Q_t^{(h)}(K_t^{(h)})^T/\sqrt{d})
\end{align}
\begin{align}
    map_{sa} = GAP(map_{sa}^{(h)})
\end{align}
This is because the VQ diffusion model \cite{gu2022vector} uses a large embedding dimension of 1024, causing the sum of softmax values to decrease too much depending on the location. If we only multiply the persistent weight by the low-frequency location without the process shown in lines 5 and 10 above, the tokens corresponding to the low-frequency location will be more likely to remain probabilistically during the sampling process, which can unintentionally hinder the generation of the object. Therefore, multiplying weight "a" in high-frequency locations helps maintain the ratio of low and high-frequency tokens while giving the effect of setting the persistent weight to 1.

\subsection{Hyper-parameter setting}
In our experiment, we set the self-attention threshold ($\phi$) to 0.45 and the persistent weight to 15 for hyper-parameter setting. For our learnable TCTS model, we used same architecture in VQ diffusion \cite{gu2022vector}, but  we reduced the number of layers from 19 to 16 when using the COCO dataset, and we reduced the hidden embedding size from 1024 to 512 when using the CUB dataset.

\section{Additional Samples}
\subsection{Over-simplification samples}
\begin{figure*}[ht]
\centering
\vspace{0.2in}
    \includegraphics[width=1.0\linewidth]{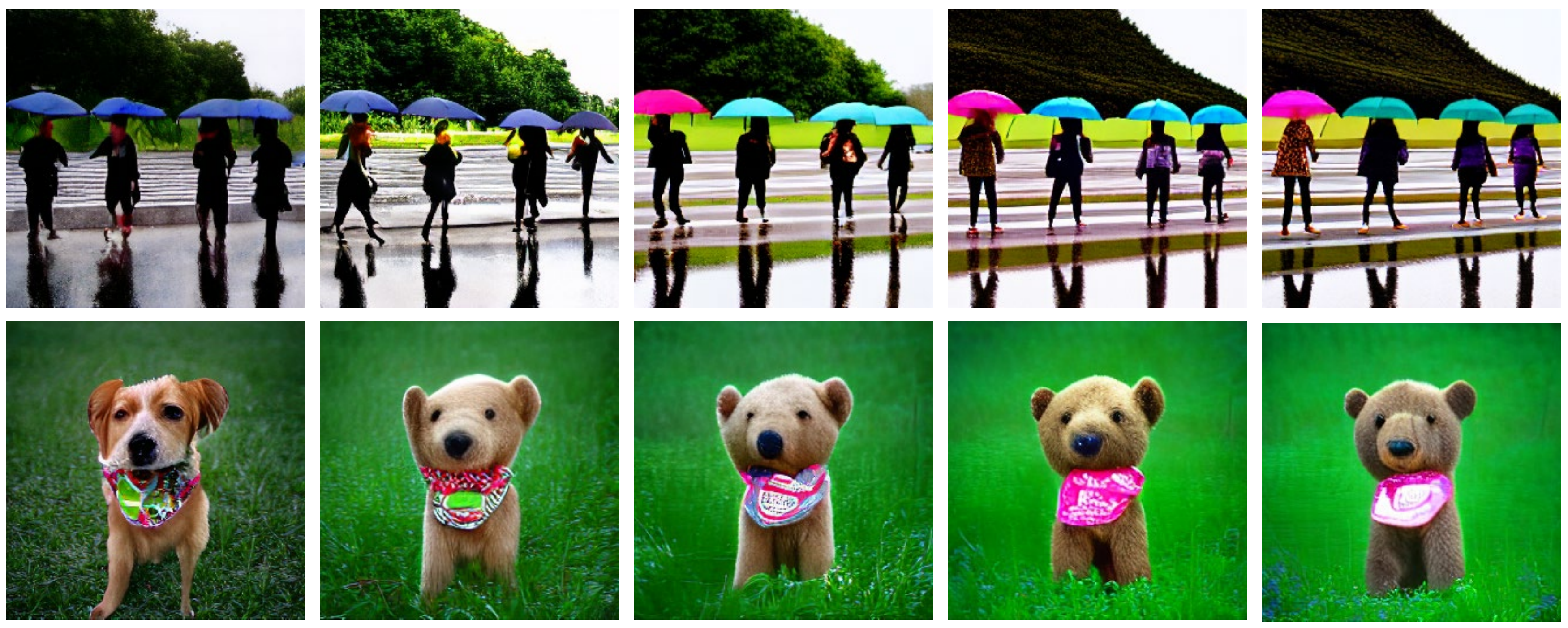}
\vspace{-0.2in}
    \caption{\small \textbf{Over-simplification samples by revocable schedule with long inference steps.} \textbf{Top}: Image generation process with a long inference steps (100 steps). \textbf{Bottom}: Mask-free object editing with long steps, "A \textbf{dog} with his tongue hanging out in a field" to "A \textbf{bear} with his tongue hanging out in a field"}
    \label{fig:appn_simp}
\vspace{-0in}
\end{figure*}

\subsection{Mask-free object editing}
\begin{figure*}[ht]
\centering
\vspace{0.1in}
    \includegraphics[width=1.0\linewidth]{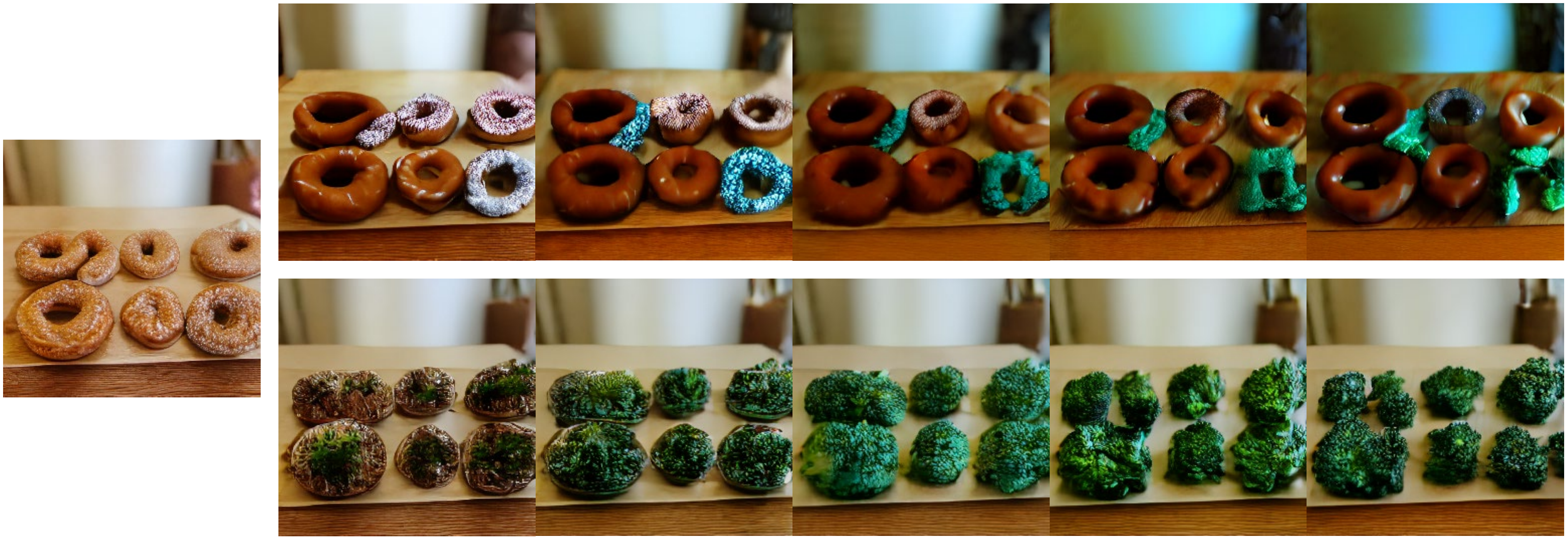}
\vspace{-0.2in}
    \caption{\small \textbf{Mask-free object editing samples with and without cross-attention map weighted sampling}. Starting from the image on the left, the result images every 20 steps of editing with 30\% masking ratio. \textbf{Top}: Failure case without weighted sampling, \textbf{Bottom}: Results with weighted sampling.}
    \label{fig:appn_mask_free}
\vspace{-0.2in}
\end{figure*}
In mask-free object editing, it is challenging to change large objects, converting donuts into broccolis for example, with a low masking ratio. This is because the distributions of token for each objects are entirely different, and even if some parts are masked, the surrounding tokens of the original object can still influence the outcome. Additionally, resampling of the whole image can lead to significant changes in unnecessary parts, such as the background. To address this issue, we propose using a cross-attention map to give more weight to sampling around the object of interest, minimizing unnecessary resampling of backgrounds, which leads to easy editing of the object.
\subsection{Image Refinement}

\begin{figure*}[ht]
\centering
\vspace{0.2in}
    \includegraphics[width=1\linewidth]{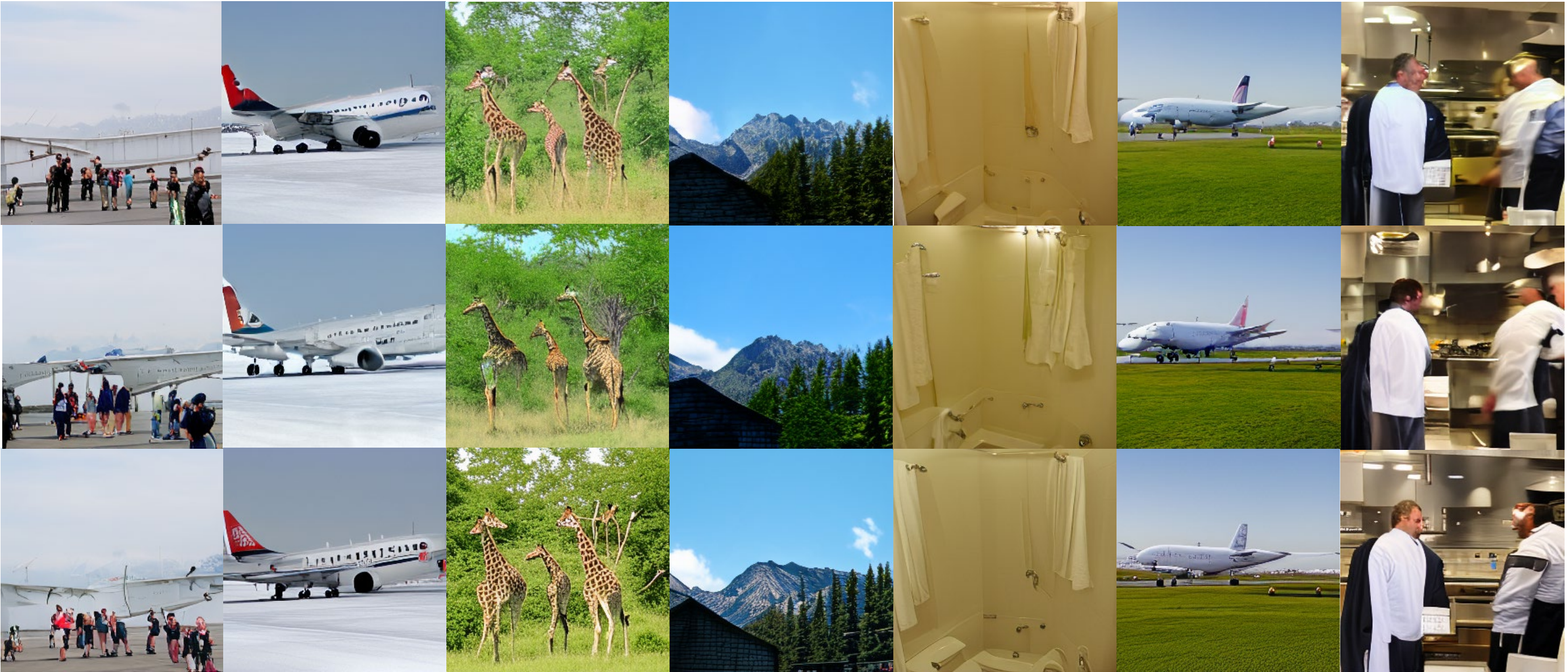}
\vspace{-0.1in}
    \caption{\small \textbf{Random image samples with additional refinement steps.} \textbf{Top}: Original images with \emph{uniform sampling} in 16 steps, \textbf{Middle}: Refined images with random noise, \textbf{Bottom}: Refined images with TCTS.}
    \label{fig:appn_refine_1}
\vspace{-0.2in}
\end{figure*} 
\begin{figure*}[ht]
\centering
\vspace{0.2in}
    \includegraphics[width=0.75\linewidth]{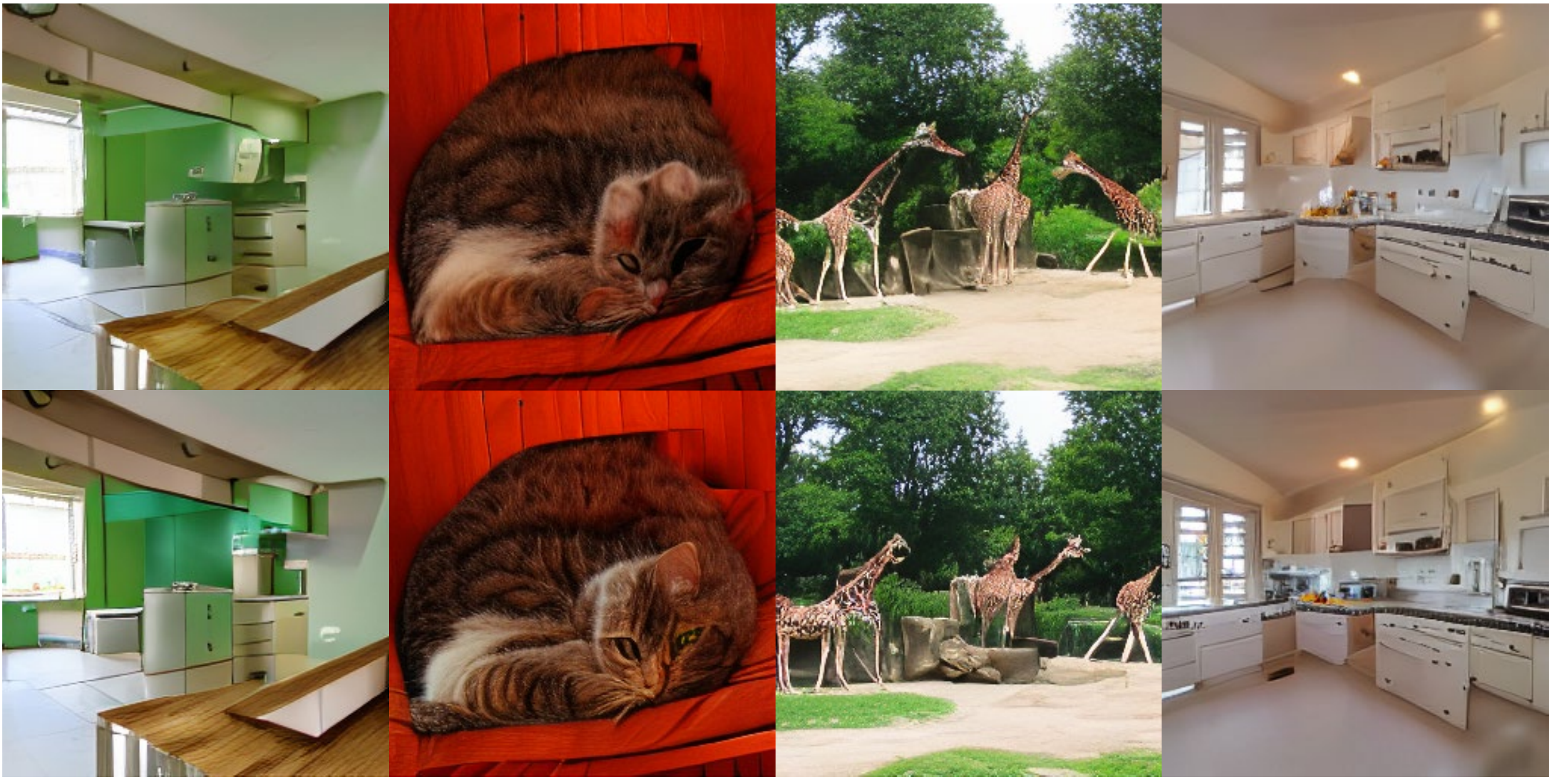}
\vspace{-0.1in}
    \caption{\small \textbf{Random image samples by refinement with masking lowest-scoring tokens.} \textbf{Top}: Original images with \emph{uniform sampling} in 16 steps, \textbf{Bottom}: Refined images with masking TCTS lowest-scoring tokens. As mentioned earlier, since it is regenerated using the same uniform sampling method, it is difficult to confirm a noticeable improvement in image quality. However, there was an improvement in performance in terms of FID and MID.}
    \label{fig:appn_refine_2}
\vspace{-0.2in}
\end{figure*}

\begin{figure*}[ht]
\centering
\vspace{0.2in}
    \includegraphics[width=1.\linewidth]{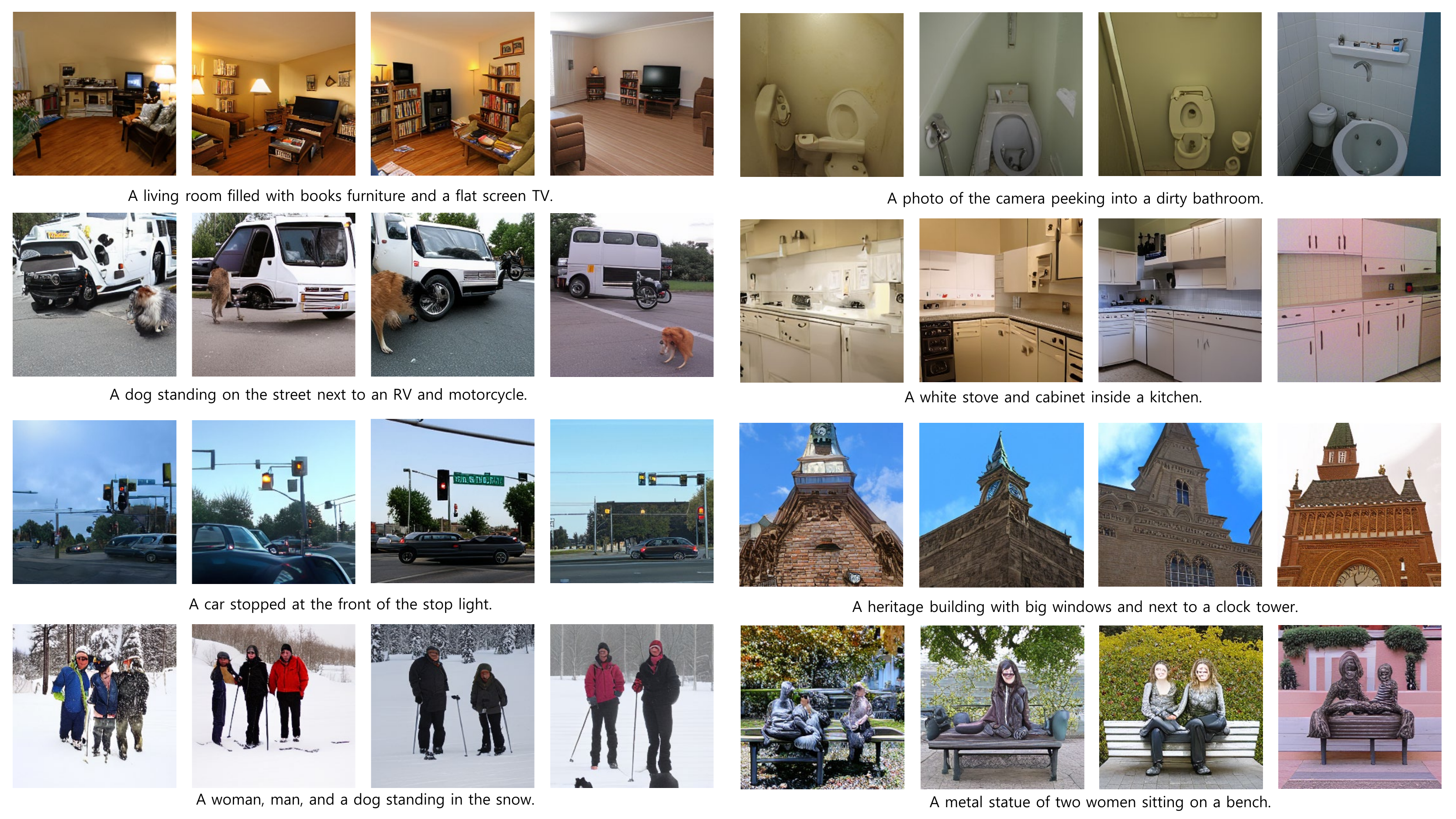}
\vspace{-0.1in}
    \caption{\small \textbf{Samples generated with TCTS. Four images are generated for each text in 8 steps, 16 steps, 25 steps, 50 steps.}}
    \label{fig:appn_step_sample}
\vspace{-0.2in}
\end{figure*} 
\begin{figure*}[ht]
\centering
\vspace{0.2in}
    \includegraphics[width=1.\linewidth]{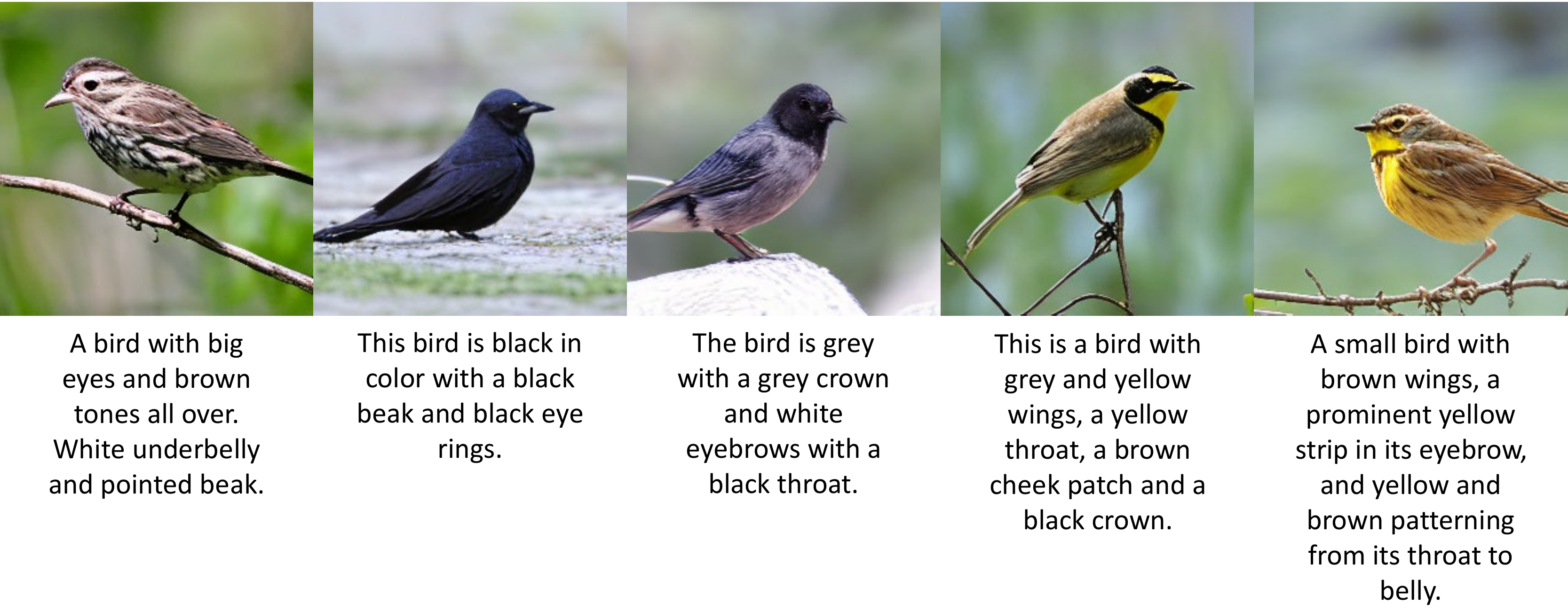}
\vspace{-0.3in}
    \caption{\small \textbf{Samples generated by TCTS with FAS in 16 steps.}}
    \label{fig:appn_cub_sample}
\vspace{-0.2in}
\end{figure*}

\section{Further Analysis on Results}
\subsection{Performance graph over time}
\begin{figure}[ht]
\centering
\vspace{0.2in}
    \includegraphics[width=1.0\linewidth]{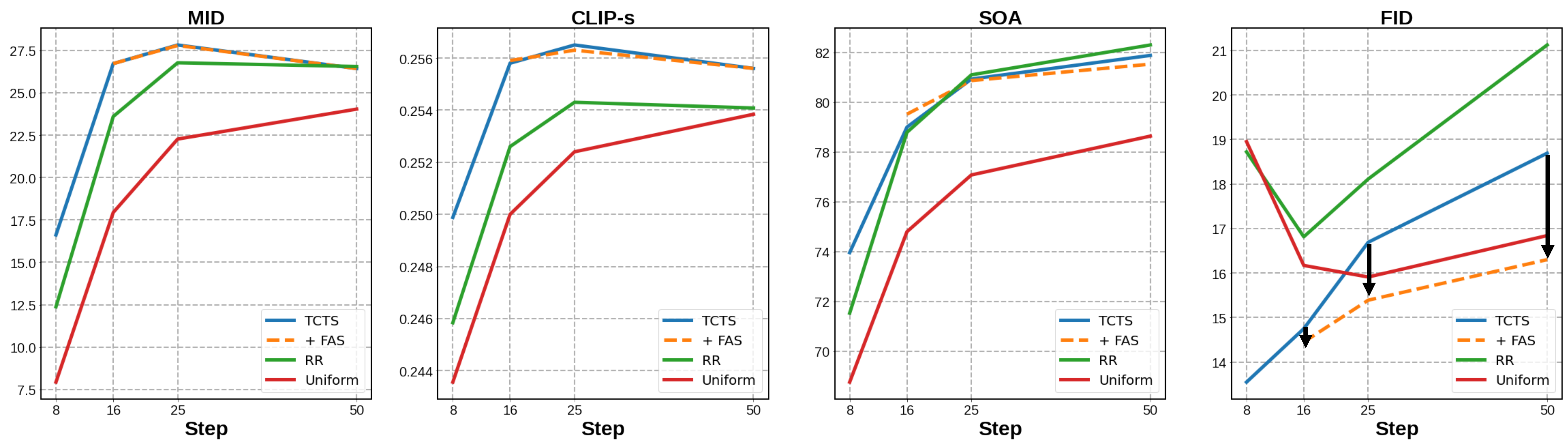}
\vspace{-0.2in}
    \caption{\small \textbf{Performance comparison of each method at different steps.} In our experiments, we fixed classifier-free guidance to 5. When we use FAS method, it was possible to lower the FID score while maintaining text alignment.}
    \label{fig:appn_step}
\vspace{-0in}
\end{figure}
\fref{fig:appn_step} demonstrates that TCTS outperforms other baseline methods in terms of MID and CLIP scores, which are relevant metrics for text. RR, TCTS, and FAS exhibit superior performance in SOA than Uniform, thus providing evidence for our analysis that revocable methods offer more opportunities for recovery, leading to the regeneration of missing objects. Furthermore, the figure illustrates the impact of FAS, which significantly enhances TCTS's FID while preserving the alignment between the image and text.
\begin{figure}[ht]
\centering
\vspace{0.2in}
    \includegraphics[width=1.0\linewidth]{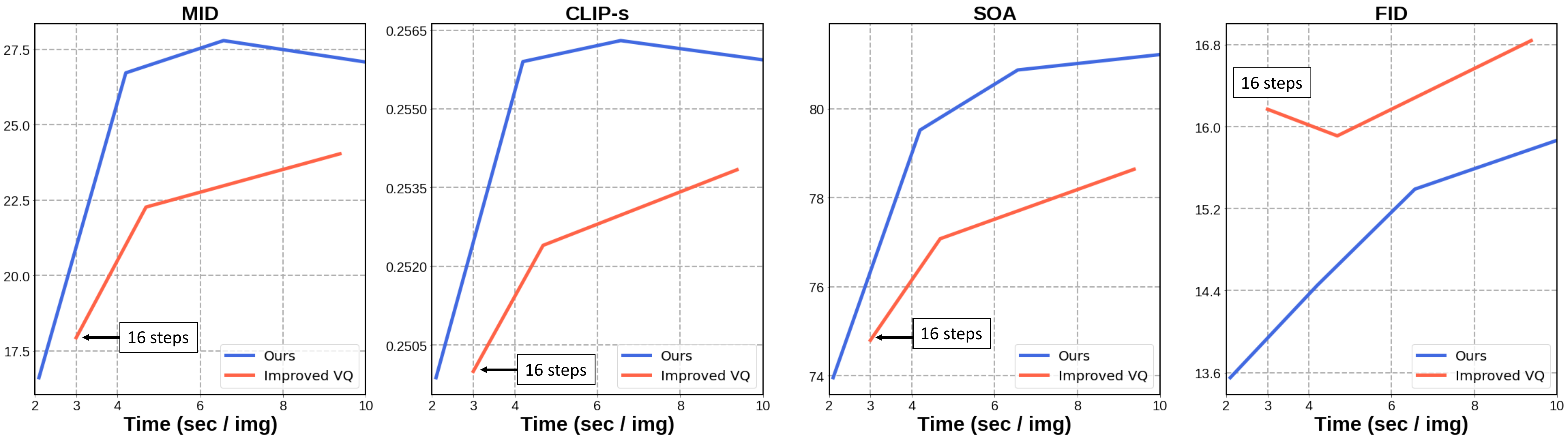}
\vspace{-0.2in}
    \caption{\small \textbf{Comparison of our model and the baseline in performance over generation time.} In our experiments, we fixed classifier-free guidance to 5.}
    \label{fig:appn_time}
\vspace{-0in}
\end{figure}

We evaluated the speed and quality of our final model against the baseline, Improved VQ-Diffusion \cite{tang2022improved}. The baseline method requires three seconds per image to generate an image in 16 steps. As shown in \fref{fig:appn_time}, our model surpasses the baseline in all metrics while maintaining the same generation time.

\end{document}